\title{IIT KGP BTP}
\author{Dvij Kalaria}
\date{February 2021}

\documentclass[12pt]{report}
\usepackage[utf8]{inputenc}

\usepackage{amsmath}
\usepackage{lipsum}
\usepackage{graphicx}
\usepackage{caption}
\usepackage{subcaption}
\usepackage{longtable}
\graphicspath{{figures/}}
\usepackage[a4paper,width=160mm,top=25mm,bottom=25mm]{geometry}
\usepackage{titlesec}
\usepackage{subcaption}
\usepackage{schemabloc,tikz}
\usepackage{amsfonts,epsfig,array,multirow,graphicx,amsmath,amsthm,ltablex,tabularx,setspace,arydshln,amssymb,multirow}
\usetikzlibrary{circuits, arrows}

\usepackage[utf8]{inputenc}
\usepackage{amsmath}
\usepackage{amsfonts}
\usepackage{amssymb}
\usepackage{bbm}
\usepackage{mathtools}
\usepackage{textcomp}
\usepackage{stackengine}
\usepackage{booktabs}
\usepackage{longtable}
\usepackage{multirow}
\usepackage{graphicx}
\usepackage{subcaption}
\usepackage{algorithmicx}
\usepackage[ruled]{algorithm}
\usepackage[noend]{algpseudocode}
\usepackage{textcomp}
\usepackage{paralist}

\def\degree{Bachelor of Technology}

\usepackage{biblatex}
\begin{document}

%Front Matter
\thispagestyle{empty}
\begin{center}
\vspace*{0.5cm}
    { \Large {\bfseries {BTP report}} \par}

% \vspace{0.2\baselineskip}
%     {\bf \degree \par}
% \vspace{0.2\baselineskip}
%     {\textit{in} \par}
% \vspace{0.2\baselineskip}
%     {\Large \bf \mydegree \par} 
% \vspace{\baselineskip}
%     {\textit{by} \par}
\vspace{\baselineskip}
    {{\Large {\bf Dvij Kalaria \\ 18CS10018}} \par}
\vspace{1.5\baselineskip}
    {\large Under the supervision of \par}
\vspace{0.3\baselineskip}
    {{\Large \bf Prof. Partha Pratim Chakrabarti \& Prof. Aritra Hazra} \par}
\vspace{\baselineskip}
    {\begin{figure}[!h] 
	\centering
	\includegraphics[width=60mm]{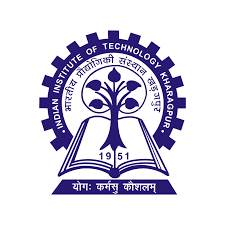} 
     \end{figure}
    }
    \vspace{1.5\baselineskip}
    {\large \MakeUppercase{Department of Computer Science \& Engineering} \par}
\vspace*{2ex}
    {\large \uppercase{Indian Institute of Technology Kharagpur} \par}
\end{center}
 \pagenumbering{roman}
% \input{FrontMatter/Acknowledgement}
% \addcontentsline{toc}{chapter}{Acknowlegement}
% \input{FrontMatter/Abstract}
% \addcontentsline{toc}{chapter}{Abstract}

\tableofcontents
% \listoffigures
% \listoftables

%Chapters
\clearpage
 \pagenumbering{arabic}
\chapter{Introduction}

\section{Adversaries}

\hspace{10mm}A deep learning model is trained on certain training examples for various tasks such as classification, regression etc. By training, weights are adjusted such that the model performs the task well not only on training examples judged by a certain metric but has an excellent ability to generalize on other unseen examples as well which are typically called the test data. Despite the huge success of machine learning models on a wide range of tasks, security has received a lot less attention along the years. Robustness along various potential cyber attacks also should be a metric for the accuracy of the machine learning models. These cyber attacks can potentially lead to a variety of negative impacts in the real world sensitive applications for which machine learning is used such as medical and transportation systems. Hence, it is a necessity to secure the system from such attacks.
Int this report, I focus on a class of these cyber attacks called the adversarial attacks in which the original input sample is modified by small perturbations such that they still look visually the same to human beings but the machine learning models are fooled by such inputs. These modified input examples are called adversarial examples or adversaries. This has become an active research topic since the first adversarial example was suspected by Amodei et al. \cite{amodei2016concrete} 

\section{Adversarial attacks}
\hspace{10mm}These are a special category of potential cyber attacks that can be made in which the attacker modifies the input examples such that they visually look the same ad appealing to the human being but the machine learning model gets fooled into mis-classifying it or mis-predicting the output value by a higher margin that too with very high probability. These attacks can be divided into white box and black box attacks. In the white box attacks, the attacker has the access to the machine learning model and its weights. In the following case, the attacks are mostly derived from using the model weights for back propagation and getting a small perturbation along the gradient direction to get the desired change in output. The second class of attacks are called black box attacks in which the attacker doesn't have access to the model but only the sample inputs and outputs to the model. These attacks typically use some characteristic properties to modify input that drives it considerably away from the input distribution or use the input and output data to construct the approximate parameters of the model and then devise a white box attack for it. 

\section{Adversarial Attack Models and Methods} \label{sec:background}
% \todo[inline]{Heading shall be "Adversarial Attack Models and Methods" and push `Datasets' and `Models' at Experimental Section, and `CVAE' to Proposed Work section next.}
For a test example $X$, an attacking method tries to find a perturbation, $\Delta X$ such that $|\Delta X|_k \leq \epsilon_{atk}$ where $\epsilon_{atk}$ is the perturbation threshold and $k$ is the appropriate order, generally selected as $2$ or $\infty$ so that the newly formed perturbed image, $X_{adv} = X + \Delta X$. Here, each pixel in the image is represented by the ${\tt \langle R,G,B \rangle}$ tuple, where ${\tt R,G,B} \in [0, 1]$. In this paper, we consider only white-box attacks, i.e. the attack methods which have access to the weights of the target classifier model. However, we believe that our method should work much better for black-box attacks as they need more perturbation to attack and hence should be more easily detected by our framework. For generating the attacks, we use the library by \cite{li2020deeprobust}. 

\subsection{Random Perturbation (RANDOM)}
Random perturbations are simply unbiased random values added to each pixel ranging in between $-\epsilon_{atk}$ to $\epsilon_{atk}$. Formally, the randomly perturbed image is given by,
\begin{equation}
X_{rand} = X + \mathcal{U}(-\epsilon_{atk},\epsilon_{atk})
\end{equation}
where, $\mathcal{U}(a,b)$ denote a continuous uniform distribution in the range $[a,b]$.

\subsection{Fast Gradient Sign Method (FGSM)}
Earlier works by~\cite{goodfellow2014explaining} introduced the generation of malicious biased perturbations at each pixel of the input image in the direction of the loss gradient $\Delta_X L(X,y)$, where $L(X,y)$ is the loss function with which the target classifier model was trained. Formally, the adversarial examples with with $l_\infty$ norm for $\epsilon_{atk}$ are computed by,
\begin{equation}
X_{adv} = X + \epsilon_{atk} . sign(\Delta_X L(X,y))
\end{equation}
FGSM perturbations with $l_2$ norm on attack bound are calculated as,
\begin{equation}
X_{adv} = X + \epsilon_{atk} . \frac{\Delta_X L(X,y)}{|\Delta_X L(X,y)|_2}
\end{equation}

\subsection{Projected Gradient Descent (PGD)}
Earlier works by~\cite{Kurakin2017AdversarialML} propose a simple variant of the FGSM method by applying it multiple times with a rather smaller step size than $\epsilon_{atk}$. However, as we need the overall perturbation after all the iterations to be within $\epsilon_{atk}$-ball of $X$, we clip the modified $X$ at each step within the $\epsilon_{atk}$ ball with $l_\infty$ norm. 
\begin{subequations}
\begin{flalign}
& X_{adv,0} = X,\\
& X_{adv,n+1} = {\tt Clip}_X^{\epsilon_{atk}}\Big{\{}X_{adv,n} + \alpha.sign(\Delta_X L(X_{adv,n},y))\Big{\}}
\end{flalign}
\end{subequations}
Given $\alpha$, we take the no of iterations, $n$ to be $\lfloor \frac{2 \epsilon_{atk}}{\alpha}+2 \rfloor$. This attacking method has also been named as Basic Iterative Method (BIM) in some works.

\subsection{Carlini-Wagner (CW) Method}
\cite{carlini2017towards} proposed a more sophisticated way of generating adversarial examples by solving an optimization objective as shown in Equation~\ref{carlini_eq}. Value of $c$ is chosen by an efficient binary search. We use the same parameters as set in \cite{li2020deeprobust} to make the attack.
\begin{equation} \label{carlini_eq}
X_{adv} = {\tt Clip}_X^{\epsilon_{atk}}\Big{\{}\min\limits_{\epsilon} \left\Vert\epsilon\right\Vert_2 + c . f(x+\epsilon)\Big{\}}
\end{equation}

\subsection{DeepFool method}
DeepFool \cite{moosavidezfooli2016deepfool} is an even more sophisticated and efficient way of generating adversaries. It works by making the perturbation iteratively towards the decision boundary so as to achieve the adversary with minimum perturbation. We use the default parameters set in \cite{li2020deeprobust} to make the attack.

%%%%%%%%%%%%%% TABLE HERE %%%%%%%%%%%%%

\section{Detection of adversaries}

\hspace{10mm}There has been an active research in the direction of adversaries and the ways to avoid them. There are many statistical, as well as machine learning based algorithms which have been put forth for the systematic detection as well as fixing them so that they get classified into the right class or give the desired output. However, mostly the literature and study has been into the adversaries in classification neural networks. This report will specifically focus on the detection of the adversaries for the classification neural networks. In other words, the ocus of this report is to briefly discuss and review the state of the art methods which are used to classify the adversaries from a pool of input examples 
\chapter{Early Statistical Methods}
Some of the early works in this domain which mark the starting of research are from Hendrys \cite{hendrycks2017early} and Gimpel \cite{hendrycks2018baseline}. They investigated 3 statistical methods which can be used to characterize adversaries from non-adversaries and hence based on these statistics identify adversaries based on threshold or machine learning techniques

\section{Using PCA}
\hspace{10mm}The first method proposed was to analyze the principal components using the SGD decomposition of the co-variance matrix to get the matrix U which can be used for finding the PCA components given the input. This modified input is called the PCA whitened input. The author proposes that for the adversaries, the magnitude of the larger PCA coefficients is very less as compared to the non adversaries. Hence by putting a threshold or some other statistical adaptive measure on the variance of the last PCA coefficients, the adversaries can be identified. This was tested for the MNIST dataset. The PCA component value variations can be seen from fig 1. The threshold based technique demonstrated the identification of FGSM \cite{goodfellow2015explaining}, BIM \cite{kurakin2017adversarial} adversaries on MNIST, Tiny-ImageNet datasets when the attacker is not aware of the defense strategy  

\begin{figure}[ht]
    \centering
    \includegraphics[width=0.7\textwidth]{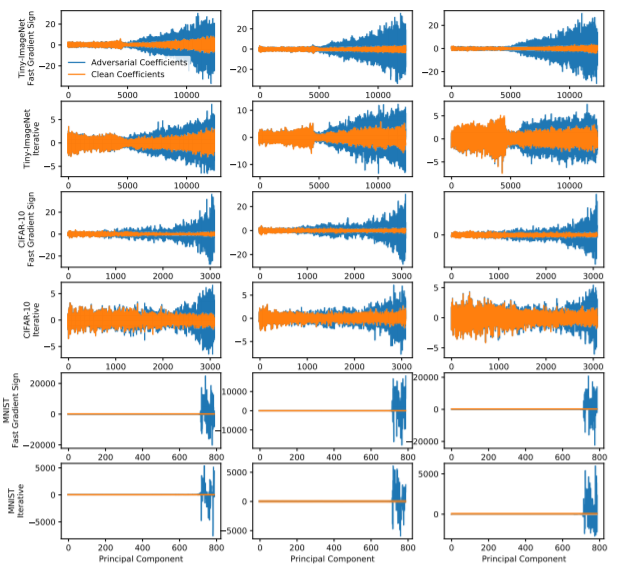}
    \caption{Magnitudes of PCA components for normal and adversarial examples}
    \label{fig1}
\end{figure}

\begin{figure}[ht]
    \centering
    \includegraphics[width=0.7\textwidth]{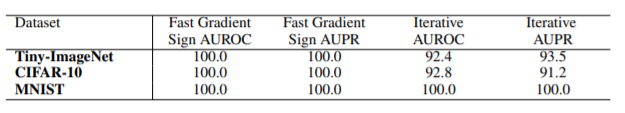}
    \caption{Percentage of adversaries correctly identified using PCA method}
    \label{table1}
\end{figure}

An extension to this work \cite{li2017adversarial} used the PCA components of deep layers instead of the input. In this work, a deep learning classifier was used for each layer of the architecture with the inputs as the PCA coefficients, max and min values and 25\%, 50\% and 75\% medians of the PCA coefficient values. For combining the result from each layer, they used a cascaded boosting. With this setup, for a sample to be identified as a non adversary, it has to be identified as non adversary at each stage. The algorithm for the following is given in fig 2.   

\begin{figure}[ht]
    \centering
    \includegraphics[width=0.7\textwidth]{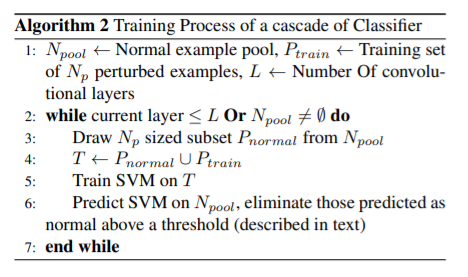}
    \caption{Algorithm for training process of a cascade classifier}
    \label{fig2}
\end{figure}

\section{Using Softmax Distribution}
\hspace{10mm}Further, Hendrycks and Gimpel \cite{hendrycks2017early} \cite{hendrycks2018baseline} proposed that the distribution after softmax on logits are different between clean and adversarial inputs, and thus can be analyzed to perform adversarial detection. The analysis measure can be done from the Kullback-Leibler divergence between uniform distribution and the softmax distribution. Then threshold-based detection or similar statistical differentiation can be made on it. It was found that the softmax distribution of normal examples are usually further away from uniform distribution compared to adversarial examples. This can be understood as a model tends to predict an input with a high confidence tend to have a uniform distribution as they essentially have equal no of training samples and thus do not differentiate on any dimension, while for adversaries generated by attack methods, do not care about the output distribution hence there is a high chance that they form a non uniform biased distribution

This method seems to only applicable to specific class of attacks that stop as soon as an input becomes adversarial such as the JSMA \cite{papernot2015limitations} , which may be predicted by the model with a low confidence score. Furthermore, evaluation on ImageNet models is needed which is currently incomplete since the softmax probability on ImageNet dataset usually is less confident due to the large number of classes. However, it is also not
clear whether or not this strategy would work against adversarial attacks that also target specific confidence score such as the C\&W attack \cite{carlini2017evaluating}

\begin{figure}[ht]
    \centering
    \includegraphics[width=0.7\textwidth]{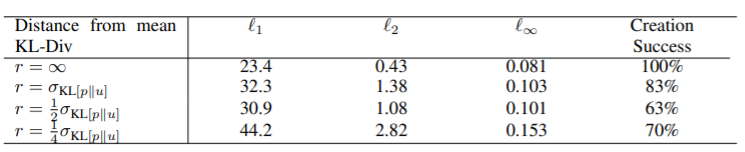}
    \caption{Constraints degrade the pathology of fooling images. Value $\sigma$ is standard deviation}
    \label{table2}
\end{figure}

\section{Using Input Reconstructions From logits}

Finally, Hendrycks and Gimpel proposed that by analyzing the input reconstructions obtained by adding an auxiliary decoder [233] to the classifier model that takes
the logits as an input, adversarial examples can be detected . The decoder and classifier are jointly trained only on clean
examples. The detection can be done by creating a detector network which takes as input the reconstruction, logits, and confidence score, and outputs a probability of an input being adversarial or not. Here, the detector network is trained on both clean and adversarial examples. They evaluated this method on MNIST model and demonstrated that this method is successful in detecting adversarial examples generated by FGSM and BIM. They showed qualitatively how reconstruction of adversarial examples are
noisier compared to normal inputs (see fig 3), allowing one to compute the differences as a way to detect adversarial
examples. The differences can be calculated using smootheness loss between input and reconstructed image.

\begin{figure}[ht]
    \centering
    \includegraphics[width=0.7\textwidth]{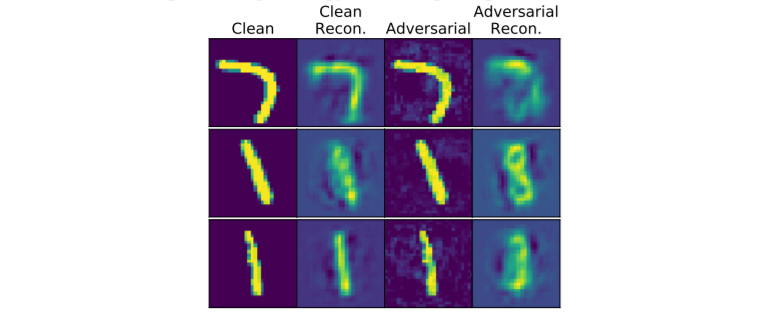}
    \caption{Adversarial image reconstruction are of lower quality then clean image reconstructions}
    \label{fig3}
\end{figure}

\section{Limitations}
\hspace{10mm}Carlini and Wagner \cite{DBLP:journals/corr/CarliniW17} in their work showed that all of these methods are not robust and ineffective against strong attacks and also inconsistent with other datasets. 

For the PCA method, they found that it can be bypassed easily if the attacker knows the defense strategy by setting a constraint on the magnitude of the magnitude of last PCA components. This can be achieved using C\&W attacks by adding this constraint to the adversary generation process. Also, they showed that the PCA method was effective on MNIST as most of the pixels are black on all images for eg the corner points but in adversaries they are changed resulting into larger magnitudes at higher PCA components. However, they argued that this was due to the aforementioned artifact in the dataset itself which made it to succeed on the MNIST dataset but however showed poor performance on the Cifar10 dataset.

The decoder and classifier networks are fully differentiable, hence can be used to pose a white box attack bypassing them. To summarize, these methods did show good result for some specific datasets against FGSM and BIM adversaries but however fail to identify complex adversary attacks like C\&W
\chapter{Network Based Methods}

\section{Adversary detector network}
\hspace{10mm}Metzen et al. \cite{metzen2017detecting} proposed augmenting a pretrained neural network with a binary detector network at each layer of the network. The detector network D is a binary classifier network that is trained to classify an example real and adversarial example. It takes the output neurons of the pre-trained network at a certain layer as an input and outputs the probability of an input being adversarial (i.e., $D(f_l(x))$ = $y_{adv}$ or $y_{clean}$, where $f_l(x)$ denotes the output of classifier f at the l-th layer).
For the training, adversary eamples are generated from the pretrained model and augmented to the dataset.
In order to account for future attacks with the assumption that attacker have access to both the classifier and detector networks, The author generated examples which specifically
attempted to fool the detector. They generated adversarial examples according to the following eqn(1) :-

$x'_{i+1} = Clip_\epsilon{x'_i + \sigma[(1-\alpha)sign(\delta_xL_{classifier}(x'_i,y)) + \alpha sign(\delta_xL_{classifier}(x'_i,y_{adv}))]}$

where $x'_0$ = x, and $\alpha$ denotes the weight factor that controls whether the attacker’s objective is to attack the
classifier or the detector, chosen randomly at every iteration. The training is first done for normal examples, the adversarial training

This method was found to be giving satisfactory results against FGSM, DeepFooland BIM attacks tested on CIFAR10 and 10-class Imagenet. Placing the detector network at different depths of the network gave different results for different attacks demonstrating the characteristic difference of the attacks. The author further proposed booging of detector networks at each of the individual layer

\begin{figure}[ht]
    \centering
    \includegraphics[width=0.7\textwidth]{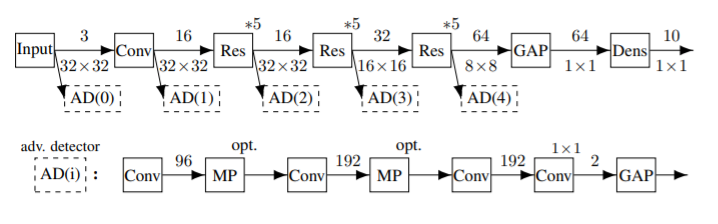}
    \caption{Detector network configuration}
    \label{fig4}
\end{figure}

Gong et al. \cite{gong2017adversarial} proposed a similar method by training a binary classifier network to differentiate between adversarial and clean examples. However, the binary classifier here is a completely separate network from the main classifier. However, rather than generating adversarial examples against the detector, they generate adversarial examples for a pretrained classifier, and augment these adversarial examples to the
original training data to train explicitly the binary classifier freezing the weights of the model.
The author observed several limitations on the generalization of this method. First, they found that the detector network is sensitive to the $\epsilon$ value used to generate FGSM and BIM adversaries, in the sense that detector trained on adversarial examples with $\epsilon_2$ cannot detect adversarial examples with $\epsilon_2$, especially when $\epsilon_2$ < $\epsilon_1$. They also discovered that training a detector was even not able to generalize on other adversaries noting that adversaries generated by different methods have different characteristics. For example, a detector trained on FGSM adversaries was found to not be able to detect JSMA adversaries  correctly.

\section{Additional Class Node}

Another similar method was proposed by Grosse et al. \cite{gong2017adversarial}, a detection method that works by augmenting a classifier network with an additional class node that represents adversarial class. Given a pretrained model, a new model with an extra class node is trained on clean examples and adversarial examples generated for the pre-trained model itself and setting the ground truth for the newly created adversaries as the new additional node.

Grosse et al. showed that this method can be used to detect adversaries generated by FGSM and JSMA robustly on MNIST. They also showed how this method can reliably
detect SBA , particularly FGSM-SBA and JSMA-SBA variants also.

\begin{figure}[ht]
    \centering
    \includegraphics[width=0.7\textwidth]{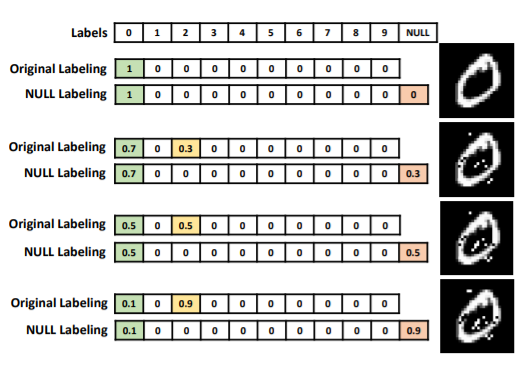}
    \caption{Example test case labels of additional class node}
    \label{fig5}
\end{figure}

Concurrent work from Hosseini et al. proposed a very similar method with slight difference in the
detail of the training procedure where the classifier is alternatively trained on clean and adversarial examples
(i.e., via adversarial training). Furthermore, the labels used for training the model were carefully assigned by performing label smoothing. Label smoothing sets a probability value to the
correct class and distributes the rest uniformly to the other classes.
Having a slightly different goal than Grosse et al. \cite{gong2017adversarial}, Hosseini et al. \cite{hosseini2017blocking} evaluated their method in
the blackbox settings and showed how their method is especially helpful to reduce the transferability rate of
adversarial examples. Although the detection method proposed by Hosseini et al. was not evaluated by Carlini and Wagner, the method appears to be similar with the method proposed by Grosse et al. Since the evaluation of this method was only done on MNIST and grayscaled GTSRB, it is doubtful whether this method will also exhibit high false positive rate when tested on CIFAR10 or other datasets with higher complexity and on other whitebox attacks. See the fig 6 for the visual diagram showing the algorithm

\begin{figure}[ht]
    \centering
    \includegraphics[width=0.7\textwidth]{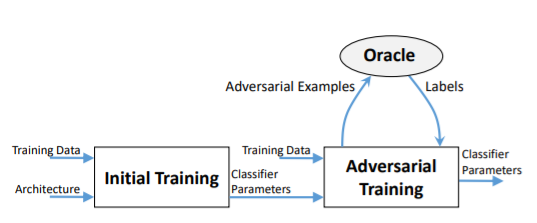}
    \caption{Training procedure}
    \label{fig6}
\end{figure}

\section{Limitations}

First of all, detector networks are differentiable, hence if their weights are accessible to the attacker in a whitebox attack, the constraint problem can be changed accordingly to bypass them. Laater, Carlini and Wagner found out that these methods have a high false positive rate against complex attacks C\&W. Also, these methods are highly sensitive to the $\epsilon$ values, if they are adversarially trained to avoid attacks for a particular value of $\epsilon$, they fail to give similar results for a slightly higher value of $\epsilon$. Also, Carlini and Wagner \cite{carlini2017evaluating} showed that similar results are not replicable on other complex datasets like CIFAR10 as on simple datasets like MNIST, hence authenticity of these methods is questionable. 
\chapter{Distribution Based ethods}

\section{Kernel Density and Bayesian Uncertainty Estimates}
\hspace{10mm}Assuming that adversarial examples are from a different distribution than the normal examples i.e. they do not lie inside the non-adversarial data manifold, Feinman et al \cite{feinman2017detecting} proposed 2 methods namely Kernel Density Estimates(KDE), where the objective is to identify a data point being in class manifold and Bayesian Uncertainty Estimates (BUE), where the objective is to find the data points in the low confidence region where the KDE is not effective. KDE works on using the logit vector distribution of clean examples to find the density estimate. The mathematical formulation of KDE is given below :- 

$KDE(x) = \frac{\sum_{x_t \in X_t} e^{\frac{-||Z(x)-Z(x_i)||^2}{\sigma ^2}}}{|X_t|}$

where Z(x) denotes the logits vector given x as the input, $X_t$ is a set of data with output class t and $\sigma$ is the bandwidth or the variance of the kernel. Note that the kernel function used is the Gaussian kernel is evaluated from the logits vecor instead of the input vector. The use of the logits is inspired from the work by Bengio et al. \cite{pmlr-v28-bengio13} and Gardner
et al. \cite{gardner2016deep}, which demonstrated how the manifold learned by the network becomes increasingly linear and flatter
which makes it easier to work with the logit space than the input space. An adversarial example x' will have
low KDE value if x' is far from the target class manifold in the logits space, and thus can be detected using threshold based or similar approach. Later, the author suggested that this method fail to work when $x'$ is near the target class
manifold.

They proposed another similar method using dropout at inference time to measure uncertainty. They mathematically quantify the uncertainty U(x) of the network by performing N stochastic passes by applying dropouts with the same setting, and using the average value of the logits as the estimate. The intuition here is that for the adversarial examples, the dependence on some specific features is quite high, hence if by dropout those deep features are dropped, a high change in value of the output is caused leading to more variance, hence a threshold can be kept on this variance to differentiate between normal and adversarial examples.
Mathematically :-

$U(x) = \frac{1}{N} \sum^{N}_{i=1}f(x)_i^Tf(x)_i - (\frac{1}{N}\sum^{N}_{i=1}f(x)_i)^T(\frac{1}{N}\sum^{N}_{i=1}f(x)_i)$

where $f(x)_i$ denotes the i-t stochastic prediction. Intuitively, this uncertainty measure should be large for adversaries. The author showed that the ratio $\frac{U(x')}{U(x)}$ was significantly more than 1 in most case when evaluated against MNIST, CIFAR10 datasets with FGSM, BIM, JSMA and also with complex adversaries from C\&W attacks. The ratio can be used to identify adversaries by putting up a simple threshold. KDE was also shown to be able to detect adversaries, but especially indicative when evaluated against a variant of BIM that stops the attack process following a fixed number of iterations. This is indicated by the ratio being less than 1 in most cases.

Based on these findings, the authors proposed a combined threshold based detection method, using
both metrics to perform adversarial detection. This is done by putting threshold values on the uncertainty
measurement and on the negative log of the kernel density estimates of a sample. The combined method
was shown to be superior than both methods individually against FGSM, BIM, JSMA, and C\&W on MNIST, CIFAR10, and SVHN.

Carlini and Wagner \cite{carlini2017evaluating} evaluated this method and concluded that the KDE method alone does not
work well on CIFAR10 and can be fooled with modified C\&W attacks by including an additional objective term defined by max(-log(KDE(x'))-$\epsilon$, 0) both in whitebox and blackbox settings. They also showed that BUE can be circumvented by C\&W attacks on the expectation values of different models sampled during dropout. However, they noted that the distortions required to generate adversarial examples that fool BUE are quite larger compared to other detection methods that they evaluated.
As a result, Carlini and Wagner \cite{carlini2017evaluating} concluded that BUE was the hardest detection method to fool compared to the other methods known to it so far. It is also relatively straightforward and easy to implement as an add-on to an existing network
architecture.

\section{Maximum Mean Discrepancy}

This particular work [x] argued that the adversaries form a different characteristic distribution as per the input samples. Hence the adversaries can be detected by finding the prior mean and variance of the cluster formed on input by the adversaries and non adversaries and using these to identify the new input sample as an adversary or non adversary. The max distance along all dimensions in the means of these 2 clusters denotes the statistical difference in the distributions of adversaries from non adversaries. This is called the Maximum Mean Discrepancy (MMD), mathematically expressed as :-

$MMD_b(F,X_1,X_2) = sup_{f\in F} (\frac{\sum_{i=1}^{N}f(x_{1i})}{N} - \frac{\sum_{i=1}^{M}f(x_{2i})}{M})$

\begin{figure}[ht]
    \centering
    \includegraphics[width=0.7\textwidth]{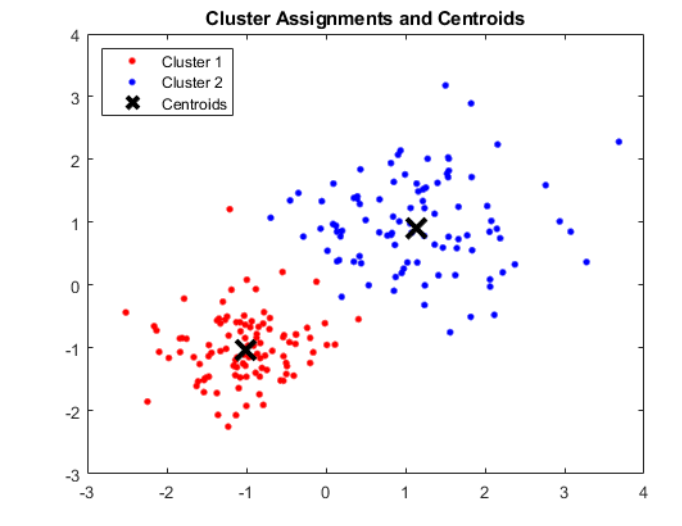}
    \caption{Clusters formed by normal samples and adversaries}
    \label{fig7}
\end{figure}

\section{Using Distribution from Deep Layers}
\hspace{10mm} PixelDefend was proposed which considered the deep layers of a pretrained network for Bayesian inference instead of input or output logit vectors. The original work is a defense technique but is related to finding a distribution for train samples and then bringing the input sample closer to that distribution for defense, hence it can be used for detection as well with the same idea using thresholding on the obtained probability value. It uses the deep representations of PixelCNN which is used for classification of images for deriving the distribution of the train samples. The probability of the sample to be non adversarial is obtained using Bayesian inference from the obtained distribution assuming it to be Gaussian. Mathematically, probability is expressed as :-

$p_{CNN}(X) = \Pi_i p_{CNN}(x_i|x_{1:(i-1)})$

To normalize the value for thresholding, Bits Per Dimension (BPD) is used which is obtained by scaling the negative of log of probability values by the dimension, as follows :-

$BPD(X) = \frac{-log p_{CNN}(X)}{I.J.K.log(2)}$

The BPD values were consistently different and characteristic for different adversaries, normal examples had less values, adversaries have higher values, strong adversaries like C\&W and DeepFool had less difference in the values, hence had higher error rates as shown in fig 4.2

\begin{figure}[ht]
    \centering
    \includegraphics[width=0.7\textwidth]{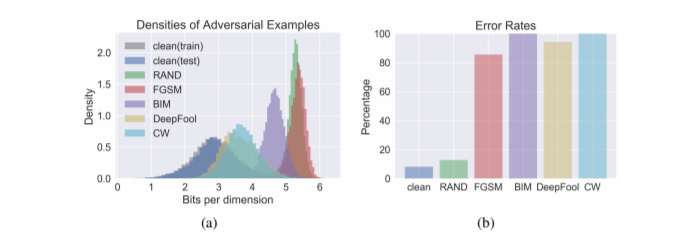}
    \caption{(a) BPD values obtained for different adversaries (b) Percentage error in results if differentiation is done on the basis of BPD threshold}
    \label{fig8}
\end{figure}

\section{Feature Squeezing}

\hspace{10mm} Xu et al. \cite{Xu_2018} argued that the with the large dimensionality of input features, yields a large attack surface. They proposed a detection strategy in which they compare the predictions of squeezed i.e. scaled down input for eg downscaled input image and unsqueezed inputs. As the name suggests, the goal of feature squeezing is to remove unnecessary features from an input by reducing the dimensionality. Two feature squeezing methods were evaluated: color bit-depth reduction and spatial
smoothing, both with local and non-local smoothing. The input is labelled as adversarial if the L1 difference (absolute difference) between the model’s prediction on squeezed and unsqueezed inputs is larger than a certain threshold value T. Different levels of squeezing can be combined by bagging i.e. taking the max of L1 difference from all of them. Figure 4.3 illustrates this method.

\begin{figure}[ht]
    \centering
    \includegraphics[width=0.7\textwidth]{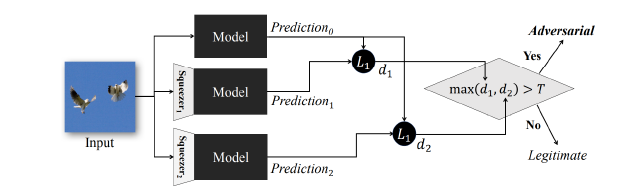}
    \caption{Procedure for squeezing model}
    \label{fig9}
\end{figure}

For color depth reduction, The author experimented with various i-bit depth where 1 $\leq$ i $\leq$ 8. For
local smoothing, median filtering was found to be more effective than Gaussian and mean smoothing. In the case of non-local smoothing, non-local means denoising was used. There is no strong motivation for this choice of squeezers as mentioned by the authors, future work on the most effective squeezers is still needed. They also suggested
using multiple feature squeezing methods as part of the detection model, since different squeezers are more
effective on certain types of adversaries than others and taking the max of the difference outputs given by each of them. Given n squeezers the output of the detector
is defined as follows :-

\[
    Detector(x)= 
\begin{cases}
    Adversarial,& \text{if } max(||f(x)-f(s_1(x)||_1,...,||f(x)-f(s_n(x)||_1)>T\\
    Benign,              & \text{otherwise}
\end{cases}
\]
where f(x) is the output vector given by the softmax layer of a deep neural network

The author also showed that feature squeezing can also be used to increase robustness along adversaries.This is similar to the work from Chuan Guo \cite{guo2018countering} which also attempted to enhance adversarial robustness through various through various input transformations. Furthermore, the author also argued that the robustness can be increased by coupling feature squeezing with adversarial training. Feature squeezing demonstrated excellent results in detecting FGSM, DeepFool, JSMA and C\&W attacks on MNIST, CIFAR10 and ImageNet datasetswhen the attacker is not aware of the strategy used. The author found that this technique does well for $L_2$ and $L_{\inf}$ based attacks and gave poor results on $L_0$ derived attacks. Intuitively, $L_0$ attacks cause perturbations that can be thought of as salt-and-pepper noise, in which median filtering has often been used to remove this type of noise. These findings support the need of using joint squeezers configuration to take into account different types of adversarial examples.

The author also considered a case where the attacker is aware of the detection model. In this setting, the attacker would typically find adversaries that fool the classifier and minimize the L1 score between squeezed and unsqueezed according to the squeezing model used. To avoid these type of whitebox attacks, the author suggested to introduce randomness for eg randomizing the threshold value as T $\pm$ rand($\delta T$).

\chapter{Some special methods}

\section{Reverse Cross Entropy}

Pang et al. \cite{pang2018robust} proposed a novel detection method by introducing a new objective function called the Reverse Cross-Entropy (RCE). For C no of classes, RCE loss is defined as :-

$L_{RCE} = -y_rlog f(x)$

Where $y_r$ is called the reversed label and is defined as :-

\[
    y_{r(i)} = 
\begin{cases}
    0,& \text{if } 
    i = y (true label)
    \\
    \frac{1}{C-1},              & \text{otherwise}
\end{cases}
\]

On training a model, to minimize $L_{RCE}$ will produce a special type of classifier called reverse classifier which will output logits such the the one with lowest value will be the predicted class. Thus, Pang et al. \cite{pang2018robust} suggested to negate the logits such
that f(x) = softmax(-Z(x)). After the model is trained, adversarial detection on the model can be performed based on threshold for some measures like Kernel density estimate (as discussed in section 4.1). They also introduced a new metric called the non maximum element (non-ME) defined as :-

$non-ME(x) = -\sum_{i\neq \hat{y}} \hat{f}(x)_i log(\hat{f}(x)_i $

where $\hat{f}(x)_i$ denotes the normalized on-max elements in f(x), For detection, the value is compared by a threshold T. If the value is less than threshold, it is classified as an adversary else as a normal valid example

This method was demonstrated to be robust when evaluated on MNIST and CIFAR10 datasets against FGSM, BIM/ILLCM \cite{kurakin2017adversarial}, JSMA \cite{buckman2018thermometer}, C\&W \cite{carlini2017evaluating} and modified C\&W that takes into account the use of KDE; which is referred to as MCW. Furthermore, this method was also evaluated in blackbox settings against MCW-SBA \cite{papernot2017practical} and showcased how the adversarial examples generated by SBA suffer from poor transferability. Even when adversarial
examples that fool both detector and classifier models were found, the adversarial examples usually exhibit larger distortions which are visually perceptible to a human.

Finally, the author argued that the use of RCE not only allows one to perform adversarial detection, but also increases robustness of a model in general compared to using standard cross-entropy as the objective function. They also did the comparison of t-SNE embeddings between a model trained with RCE and non-RCE objectives, and showed how the t-SNE \cite{JMLR:v9:vandermaaten08a} visualization of the model that was trained with RCE objective achieved higher separability. The use of new objective functions such as RCE which constrains the ability of attacker to generate adversarial examples making it robust seems to be an interesting research area to be explored in the future.

\section{Adversarial Examples Detection in Features Distance Spaces}

The authors for this work \cite{inbook} put forth a very interesting way to detect adversaries. The authors argued that the adversaries trace a certain path when considered in the feature space through all the deep layers of network. These paths can be used as a criteria to determine whether the input is an adversary is not. The set of all paths taken by the normal examples are used to determine the mean paths. The mean vector values are pre-calculated at each layer to get a 2D vector of the mean feature vectors for each sample class. Using these means at each layer, an M dimensional vector is obtained at each layer where M is the no of classes as the L2 distance from these M mean vectors at each layer. These N distance vectors are passed through sequential neural networks like LSTM or MLP. The output of this sequential network is used to predict whether the input is an adversary or not

\begin{figure}[ht]
    \centering
    \includegraphics[width=0.7\textwidth]{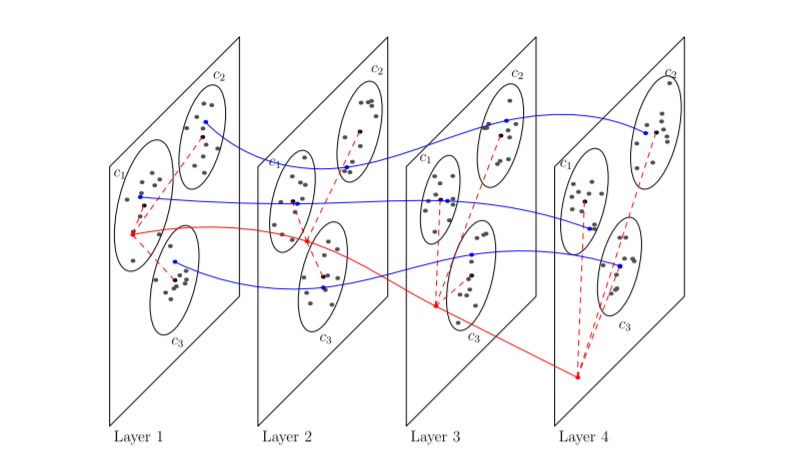}
    \caption{Intuition on distance feature paths followed by normal samples and adversaries}
    \label{fig10}
\end{figure}

Basically, this method tries to find patterns in the sequential inference through deep layers to detect the presence of adversarial example. It tries to encode sequential flow of inference within layers into training an LSTM network which outputs the probability. Hence the trained LSTM network tries to find similarities in the inference patterns which it fails to in case of adversarial examples, hence able to detect them. The basic procedure can be described as :- 

1. Calculate the principle centers for each class for each layer as the centroid of all the examples:- 
$p_c^l = \frac{\sum_{j=1}{K_c} o_{c,j}^l}{K_c}$

2. For the input, derive C neuron embedding for each layer by taking L2 distance from each of the principal centre :-

$e^l = (d(o^l,p_1^l),d(o^l,p_2^l).....d(o^l,p_C^l))$

3. Feed the L layer embedding to train a time series network like LSTM to find patterns in the sequential embedding

\begin{figure}[ht]
    \centering
    \includegraphics[width=0.7\textwidth]{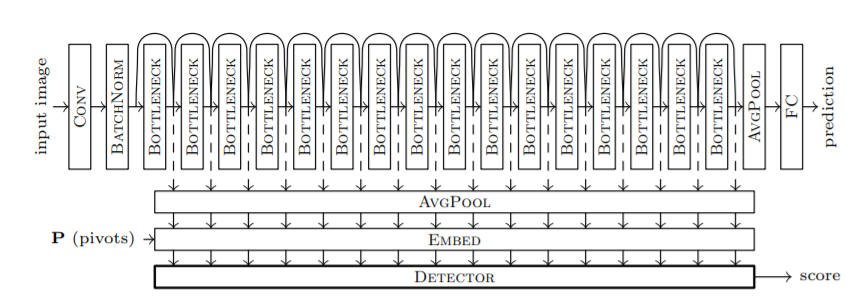}
    \caption{Network architecture}
    \label{fig11}
\end{figure}

The model demonstrated excellent results on large no of adversary attacks including L-BFGS, FGSM, BIM, PGD, MI-FGSM. Also it was observed that using LSTM yielded better results than MLP due to the inherent sequential nature of the patterns which are better recorded by the LSTM. True Positive Rate (TPR) and False Positive Rate (FPR) ratio results can be seen from fig 5.3

\begin{figure}[ht]
    \centering
    \includegraphics[width=0.5\textwidth]{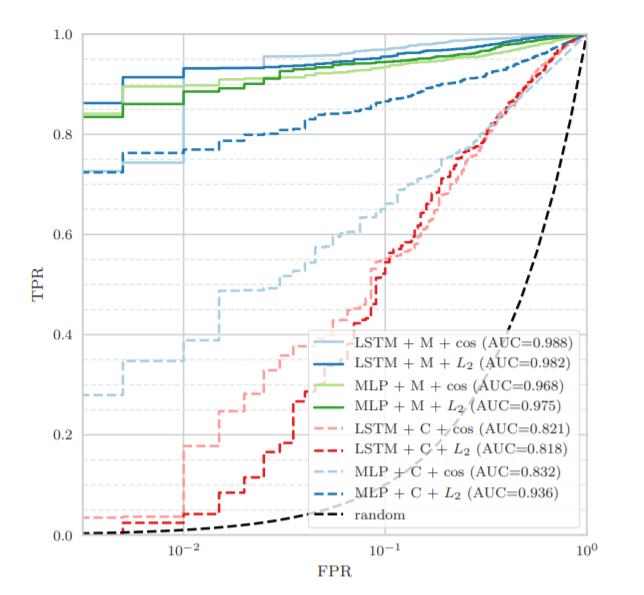}
    \caption{True positive and false positive rates for adversaries and normal samples}
    \label{fig12}
\end{figure}

\chapter{Use of conditional Variational AutoEncoder (CVAE)}

\section{Introduction} \label{sec:introduction}

Many recent works have presented effective ways in which adversarial attacks can be avoided as we discussed in earlier chapters. To re-iterate, adversarial attacks can be classified into whitebox and blackbox attacks. White-box attacks~\cite{akhtar2018threat} assume access to the neural network weights and architecture, which are used for classification, and thereby specifically targeted to fool the neural network. Hence, they are more accurate than blackbox attacks~\cite{akhtar2018threat} which do not assume access the model parameters. Methods for detection of adversarial attacks can be broadly categorized as -- (i) statistical methods, (ii) network based methods, and (iii) distribution based methods. Statistical methods~\cite{hendrycks2016early} \cite{li2017adversarial} focus on exploiting certain characteristics of the input images or the final logistic-unit layer of the classifier network and try to identify adversaries through their statistical inference. A certain drawback of such methods as pointed by~\cite{carlini2017towards} is that the statistics derived may be dataset specific and same techniques are not generalized across other datasets and also fail against strong attacks like CW-attack. Network based methods~\cite{metzen2017detecting} \cite{gong2017adversarial} aim at specifically training a binary classification neural network to identify the adversaries. These methods are restricted since they do not generalize well across unknown attacks on which these networks are not trained, also they are sensitive to change with the amount of perturbation values such that a small increase in these values makes this attacks unsuccessful. Also, potential whitebox attacks can be designed as shown by~\cite{carlini2017towards} which fool both the detection network as well as the adversary classifier networks. Distribution based methods~\cite{feinman2017detecting} \cite{gao2021maximum} \cite{song2017pixeldefend} \cite{xu2017feature} \cite{jha2018detecting} aim at finding the probability distribution from the clean examples and try to find the probability of the input example to quantify how much they fall within the same distribution. However, some of the methods do not guarantee robust separation of randomly perturbed and adversarial perturbed images. Hence there is a high chance that all these methods tend to get confused with random noises in the image, treating them as adversaries.

To overcome this drawback so that the learned models are robust with respect to both adversarial perturbations as well as sensitivity to random noises, we propose the use of Conditional Variational AutoEncoder (CVAE) trained over a clean image set. At the time of inference, we empirically establish that the input example falls within a low probability region of the clean examples of the predicted class from the target classifier network. It is important to note here that, this method uses both the input image as well as the predicted class to detect whether the input is an adversary as opposed to some distribution based methods which use only the distribution from the input images. On the contrary, random perturbations activate the target classifier network in such a way that the predicted output class matches with the actual class of the input image and hence it falls within the high probability region. Thus, it is empirically shown that our method does not confuse random noise with adversarial noises. Moreover, we show how our method is robust towards special attacks which have access to both the network weights of CVAE as well as the target classifier networks where many network based methods falter. Further, we show that to eventually fool our method, we may need larger perturbations which becomes visually perceptible to the human eye. The experimental results shown over MNIST and CIFAR-10 datasets present the working of our proposal.

In particular, the primary contributions made by our work is as follows.
\begin{compactenum}[(a)]
 \item We propose a framework based on CVAE to detect the possibility of adversarial attacks.

 \item We leverage distribution based methods to effectively differentiate between randomly perturbed and adversarially perturbed images.

 \item We devise techniques to robustly detect specially targeted BIM-attacks~\cite{metzen2017detecting} using our proposed framework.
 
%  \item We show the efficacy of our approach using MNIST and CIFAR-10 image datasets and compare the capabilities with earlier methods.
\end{compactenum}
To the best of our knowledge, this is the first work which leverages use of Variational AutoEncoder architecture for detecting adversaries as well as aptly differentiates noise from adversaries to effectively safeguard learned models against adversarial attacks.
%We believe that the proposed framework can be effectively used in safety-critical practical domains like automotive and healthcare to safeguard them against adversarial attacks.

%The rest of this paper is organized as follows. Section~\ref{sec:background} presents the adversarial attack model and existing methods for their detection. Section~\ref{sec:method} describes our proposed framework and methodology of using CVAE to detect adversarial attacks. Section~\ref{sec:experiment} showcases the experimental results and comparison with other works obtained over MNIST and CIFAR-10 datasets. In Section~\ref{sec:literature}, we also a review of the related works carried out in this domain. Finally, we conclude the paper in Section~\ref{sec:conclusion}.

% TO CHANGE

\section{Proposed Framework Leveraging CVAE} \label{sec:method}
In this section, we present how Conditional Variational AutoEncoders (CVAE), trained over a dataset of clean images, are capable of comprehending the inherent differentiable attributes between adversaries and noisy data and separate out both using their probability distribution map.
%In this section, we present our proposed framework in details to enable this in a step-by-step matter.

\subsection{Conditional Variational AutoEncoders (CVAE)}
Variational AutoEncoder is a type of a Generative Adversarial Network (GAN) having two components as Encoder and Decoder. The input is first passed through an encoder to get the latent vector for the image. The latent vector is passed through the decoder to get the reconstructed input of the same size as the image. The encoder and decoder layers are trained with the objectives to get the reconstructed image as close to the input image as possible thus forcing to preserve most of the features of the input image in the latent vector to learn a compact representation of the image. The second objective is to get the distribution of the latent vectors for all the images close to the desired distribution. Hence, after the variational autoencoder is fully trained, decoder layer can be used to generate examples from randomly sampled latent vectors from the desired distribution with which the encoder and decoder layers were trained.

%%%%%%%%%%%%%% FIG HERE %%%%%%%%%%%%%%%
\vspace{-0.3cm}
\begin{figure}[h] 
    \centering
    \includegraphics[width=0.45\textwidth]{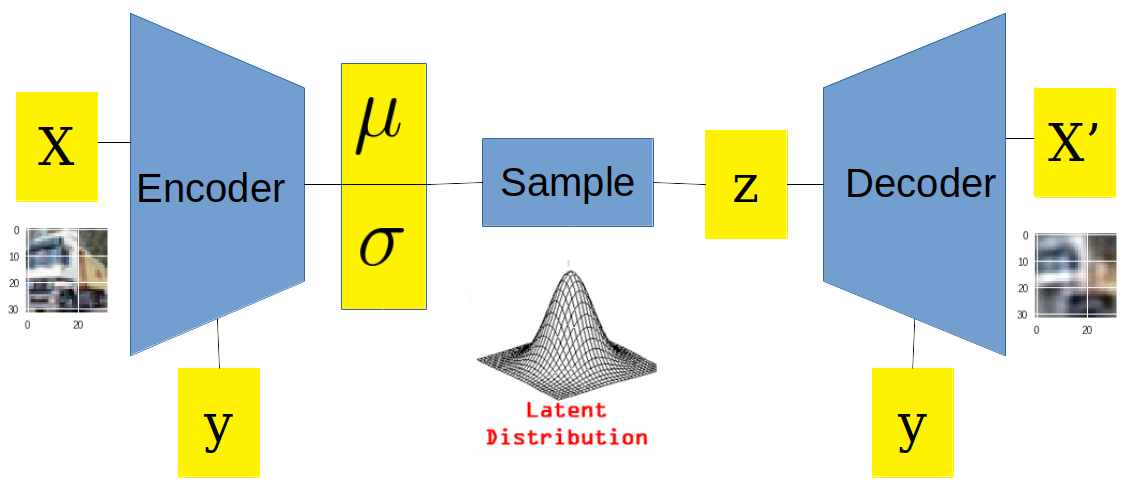}
    \caption{CVAE Model Architecture}
    \label{fig:cvae_diag}
\end{figure}
\vspace{-0.3cm}

Conditional VAE is a variation of VAE in which along with the input image, the class of the image is also passed with the input at the encoder layer and also with the latent vector before the decoder layer (refer to Figure~\ref{fig:cvae_diag}). This helps Conditional VAE to generate specific examples of a class. The loss function for CVAE is defined by Equation~\ref{eq:cvae}. The first term is the reconstruction loss which signifies how closely can the input $X$ be reconstructed given the latent vector $z$ and the output class from the target classifier network as condition, $c$. The second term of the loss function is the KL-divergence ($\mathcal{D}_{KL}$) between the desired distribution, $P(z|c)$ and the current distribution ($Q(z|X,c)$) of $z$ given input image $X$ and the condition $c$.
\begin{equation} \label{eq:cvae}
    L(X,c) = \mathbb{E} \big{[}\log P(X|z,c) \big{]} - \mathcal{D}_{KL} \big{[} Q(z|X,c)\ ||\ P(z|c) \big{]}
\end{equation}

\subsection{Training CVAE Models}
For modeling $\log P(X|z,c)$, we use the decoder neural network to output the reconstructed image, $X_{rcn}$ where we utilize the condition $c$ which is the output class of the image to get the set of parameters, $\theta(c)$ for the neural network. We calculate Binary Cross Entropy (${\tt BCE}$) loss of the reconstructed image, $X_{rcn}$ with the input image, $X$ to model $\log P(X|z,c)$. Similarly, we model $Q(z|X,c)$ with the encoder neural network which takes as input image $X$ and utilizes condition, $c$ to select model parameters, $\theta(c)$ and outputs mean, $\mu$ and log of variance, $\log \sigma^2$ as parameters assuming Gaussian distribution for the conditional distribution. We set the target distribution $P(z|c)$ as unit Gaussian distribution with mean 0 and variance 1 as $N(0,1)$. The resultant loss function would be as follows,
\begin{eqnarray}
    L(X,c) & = & {\tt BCE} \big{[} X, Decoder(X,\theta(c)) \big{]} - \nonumber\\
    & & \frac{1}{2}\Big{[}Encoder_\sigma^2(X,\theta(c))
      + Encoder_\mu^2(X,\theta(c)) \nonumber\\
    & & \qquad - 1 - \log \big{(} Encoder_\sigma^2(X,\theta(c)) \big{)} \Big{]}
\end{eqnarray}

The model architecture weights, $\theta(c)$ are a function of the condition, $c$. Hence, we learn separate weights for encoder and decoder layers of CVAE for all the classes. It implies learning different encoder and decoder for each individual class. The layers sizes are tabulated in Table~\ref{tab:cvae_arch_sizes}. We train the Encoder and Decoder layers of CVAE on clean images with their ground truth labels and use the condition as the predicted class from the target classifier network during inference.

\vspace{-0.2cm}
\begin{table}[h]
{\sf \scriptsize
\begin{center}
\begin{tabular}{|c||c|l|}
    \hline
  {\bf Attribute}  & {\bf Layer}     & {\bf Size}    \\
  \hline
  \hline
          & Conv2d      & Channels: (c, 32)\\ 
          &             & Kernel: (4,4,stride=2,padding=1)  \\
          \cline{2-3}
          & BatchNorm2d & 32 \\
          \cline{2-3}
          & Relu        &    \\
          \cline{2-3}
          & Conv2d      & Channels: (32, 64)\\ 
 Encoder  &             & Kernel: (4,4,stride=2,padding=1)  \\
          \cline{2-3}
          & BatchNorm2d & 64 \\
          \cline{2-3}
          & Relu        &    \\
          \cline{2-3}
          & Conv2d      & Channels: (64, 128)\\ 
          &             & Kernel: (4,4,stride=2,padding=1)  \\
          \cline{2-3}
          & BatchNorm2d & 128 \\
  \hline
  Mean     & Linear    & (1024,$z_{dim}$=128) \\
  \hline
  Variance & Linear    & (1024,$z_{dim}$=128)  \\
  \hline
  Project  & Linear    & ($z_{dim}$=128,1024) \\
          \cline{2-3}
          & Reshape   & (128,4,4) \\
  \hline
          & ConvTranspose2d & Channels: (128, 64)\\
          &                 & Kernel: (4,4,stride=2,padding=1)  \\
          \cline{2-3}
          & BatchNorm2d & 64 \\
          \cline{2-3}
          & Relu        &    \\
          \cline{2-3}
          & ConvTranspose2d & Channels: (64, 32)\\
 Decoder  &                 & Kernel: (4,4,stride=2,padding=1)  \\
          \cline{2-3}
          & BatchNorm2d & 64 \\
          \cline{2-3}
          & Relu        &    \\
          \cline{2-3}
          & ConvTranspose2d & Channels: (32, c)\\
          &                 & Kernel: (4,4,stride=2,padding=1)  \\
          \cline{2-3}
          & Sigmoid     &   \\
\hline
\end{tabular}
\end{center}
}
\caption{CVAE Architecture Layer Sizes. $c$ = Number of Channels in the Input Image ($c=3$ for CIIFAR-10 and $c=1$ for MNIST).}
\label{tab:cvae_arch_sizes}
\end{table}
% \vspace{-0.3cm}

% \todo[inline]{Split into two separate sections -- (a) Proposed Methodology leveraging CVAE, and (b) Experimental Results. Push description of CVAE from previous section here.

% Write an algorithm and present your approach. May be, you can draw a schematic framework to express the steps of the flow too.}

\subsection{Determining Reconstruction Errors}
\begin{figure}[ht] 
    \begin{center}
    \begin{subfigure}{.4\textwidth}
        \centering
        \includegraphics[width=\textwidth]{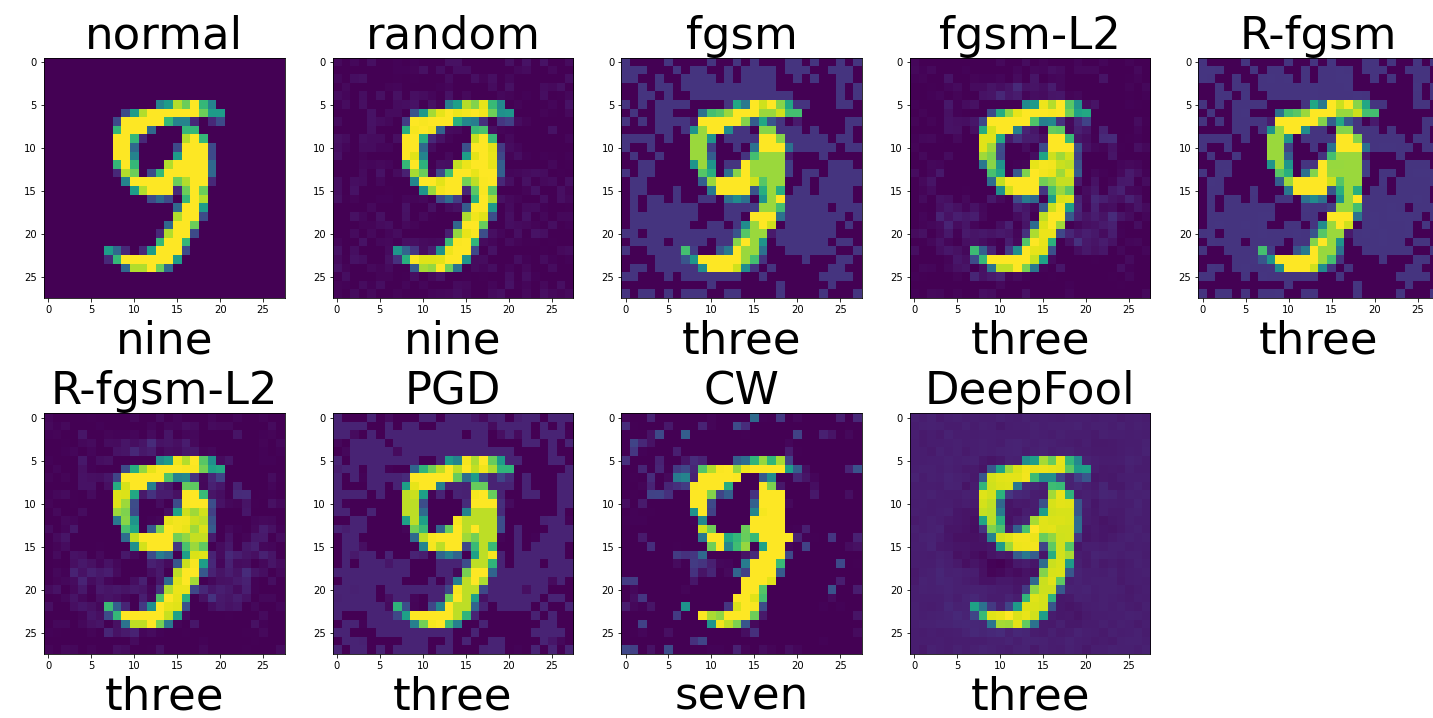}
        \caption{Input Images}
    \end{subfigure}
    \begin{subfigure}{.4\textwidth}
        \centering
        \includegraphics[width=\textwidth]{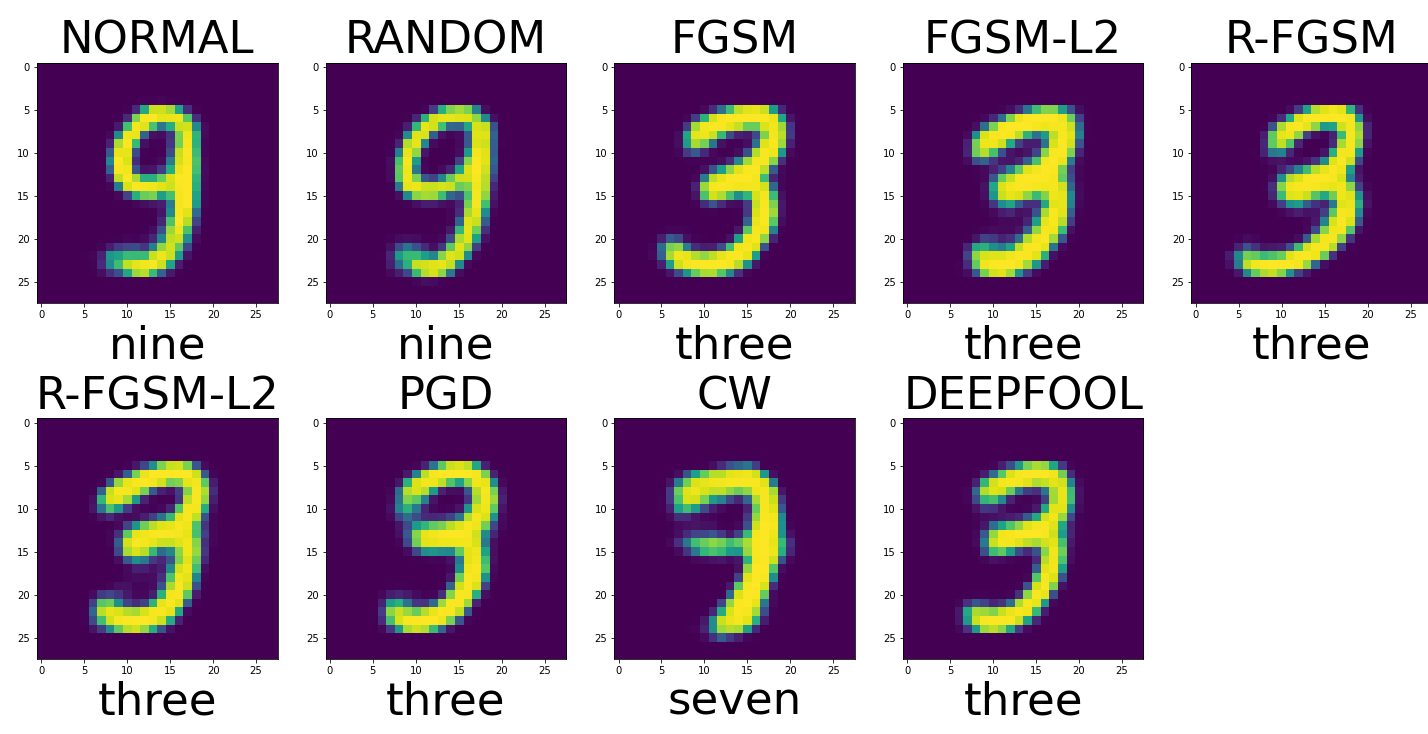}
        \caption{Reconstructed Images}
    \end{subfigure}
    \caption{Clean and Adversarial Attacked Images to CVAE from MNIST Dataset}
    \label{fig:eg_images_mnist}
    \end{center}
\end{figure}

\begin{figure}[ht] 
    \begin{center}
    \begin{subfigure}{.4\textwidth}
        \centering
        \includegraphics[width=\textwidth]{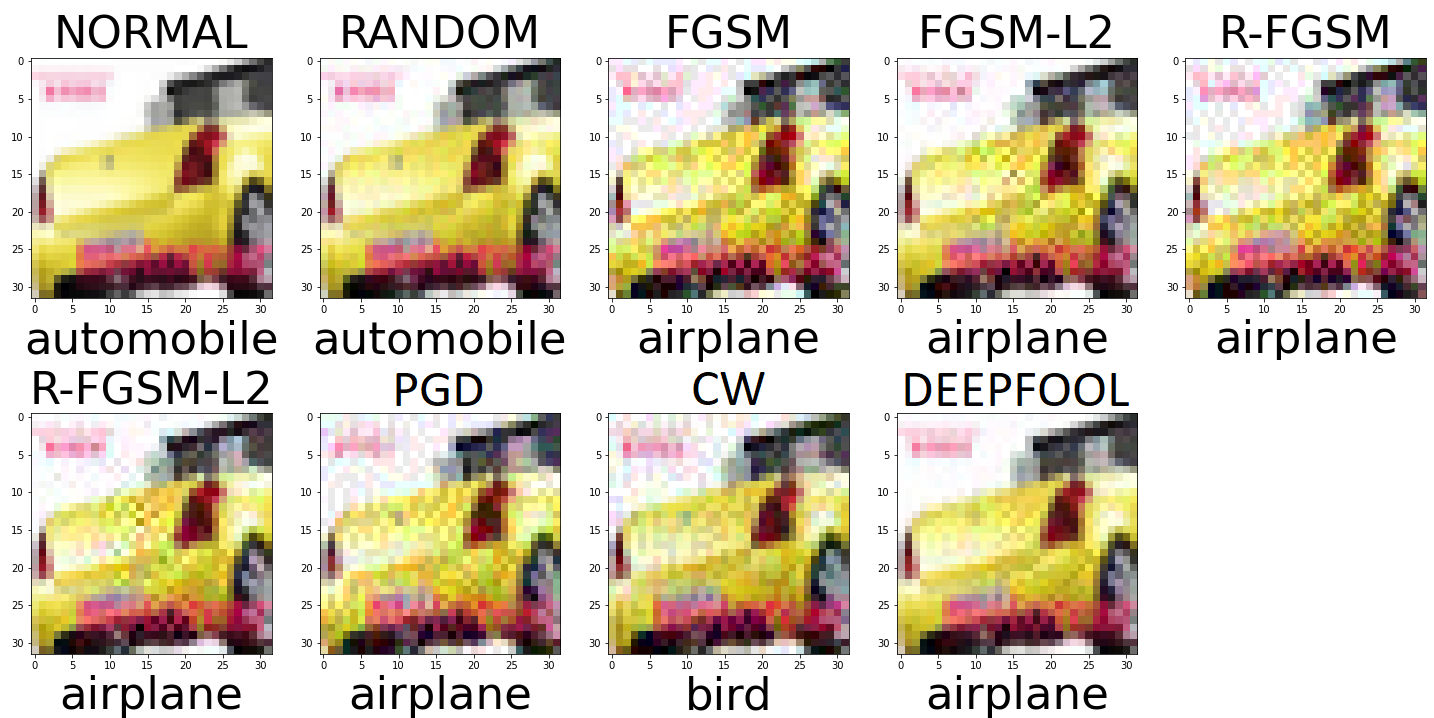}
        \caption{Input Images}
    \end{subfigure}
    \begin{subfigure}{.4\textwidth}
        \centering
        \includegraphics[width=\textwidth]{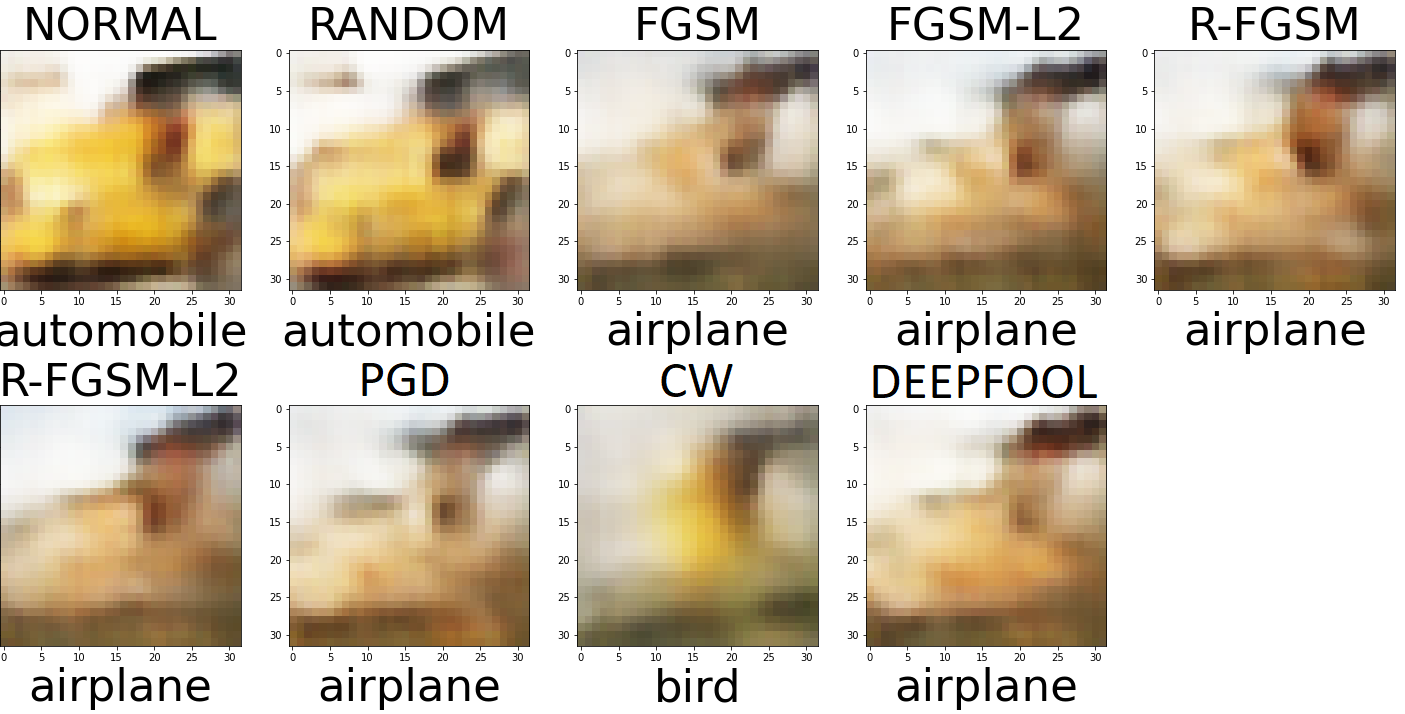}
        \caption{Reconstructed Images}
    \end{subfigure}
    \caption{Clean and Adversarial Attacked Images to CVAE from CIFAR-10 Dataset. }
    \label{fig:eg_images_cifar}
    \end{center}
\vspace{-0.5cm}
\end{figure}

Let $X$ be the input image and $y_{pred}$ be the predicted class obtained from the target classifier network. $X_{rcn, y_{pred}}$ is the reconstructed image obtained from the trained encoder and decoder networks with the condition $y_{pred}$. We define the reconstruction error or the reconstruction distance as in Equation~\ref{eq:recon}. The network architectures for encoder and decoder layers are given in Figure~\ref{fig:cvae_diag}.
\begin{equation} \label{eq:recon}
    {\tt Recon}(X,y) = (X - X_{rcn,y})^2
\end{equation}
Two pertinent points to note here are:
\begin{compactitem}
    \item For clean test examples, the reconstruction error is bound to be less since the CVAE is trained on clean train images. As the classifier gives correct class for the clean examples, the reconstruction error with the correct class of the image as input is less.
    
    \item For the adversarial examples, as they fool the classifier network, passing the malicious output class, $y_{pred}$ of the classifier network to the CVAE with the slightly perturbed input image, the reconstructed image tries to be closer to the input with class $y_{pred}$ and hence, the reconstruction error is large.
\end{compactitem}
As an example, let the clean image be a cat and its slightly perturbed image fools the classifier network to believe it is a dog. Hence, the input to the CVAE will be the slightly perturbed cat image with the class dog. Now as the encoder and decoder layers are trained to output a dog image if the class inputted is dog, the reconstructed image will try to resemble a dog but since the input is a cat image, there will be large reconstruction error. Hence, we use reconstruction error as a measure to determine if the input image is adversarial. We first train the Conditional Variational AutoEncoder (CVAE) on clean images with the ground truth class as the condition. Examples of reconstructions for clean and adversarial examples are given in Figure~\ref{fig:eg_images_mnist} and Figure~\ref{fig:eg_images_cifar}. 

%%%%%%%%%%%%%%%%%%%%% FIG HERE %%%%%%%%%%%%%%%%%
\vspace{-0.3cm}
\vspace{-0.3cm}

\subsection{Obtaining $p$-value}
% The distribution of reconstruction distances of test images are given in~\ref{fig:recons_dist}.
As already discussed, the reconstruction error is used as a basis for detection of adversaries. We first obtain the reconstruction distances for the train dataset of clean images which is expected to be similar to that of the train images. On the other hand, for the adversarial examples, as the predicted class $y$ is incorrect, the reconstruction is expected to be worse as it will be more similar to the image of class $y$ as the decoder network is trained to generate such images. Also for random images, as they do not mostly fool the classifier network, the predicted class, $y$ is expected to be correct, hence reconstruction distance is expected to be less. Besides qualitative analysis, for the quantitative measure, we use the permutation test from~\cite{EfroTibs93}. We can provide an uncertainty value for each input about whether it comes from the training distribution. Specifically, let the input $X'$ and training images $X_1, X_2, \ldots, X_N$. We first compute the reconstruction distances denoted by ${\tt Recon}(X,y)$ for all samples with the condition as the predicted class $y = {\tt Classifier}(X)$. Then, using the rank of ${\tt Recon}(X',y')$ in $\{ {\tt Recon}(X_1,y_1), {\tt Recon}(X_2,y_2), \ldots, {\tt Recon}(X_N,y_N)\}$ as our test statistic, we get,
%%%%%%%%%%%%%% EQN HERE %%%%%%%%%%%%
\begin{eqnarray}
    T & = & T(X' ; X_1, X_2, \ldots, X_N) \nonumber\\
    & = & \sum_{i=1}^N I \big{[} {\tt Recon}(X_i,y_i) \leq {\tt Recon}(X',y') \big{]}
\end{eqnarray}
Where $I[.]$ is an indicator function which returns $1$ if the condition inside brackets is true, and $0$ if false. By permutation principle, $p$-value for each sample will be,
%%%%%%%%%%%%%% EQN HERE %%%%%%%%%%%%
\begin{equation}
    p = \frac{1}{N+1} \Big{(} \sum_{i=1}^N I[T_i \leq T]+1 \Big{)}
\end{equation}
Larger $p$-value implies that the sample is more probable to be a clean example. Let $t$ be the threshold on the obtained $p$-value for the sample, hence if $p_{X,y} < t$, the sample $X$ is classified as an adversary. Algorithm~\ref{algo:adv_detect} presents the overall resulting procedure combining all above mentioned stages.

\vspace{-0.3cm}
\alglanguage{pseudocode}
\begin{algorithm}
\small
\caption{Adversarial Detection Algorithm}
\label{algo:adv_detect}
\begin{algorithmic}[1]
\Function{Detect\_Adversaries ($X_{train}, Y_{train}, X, t$)}{}
    \State recon $\gets$ ${\tt Train}(X_{train},Y_{train})$
    \State recon\_dists $\gets$ ${\tt Recon}(X_{train},Y_{train})$
    \State Adversaries $\gets$ $\phi$
    \For{$x$ in $X$}
        \State $y_{pred}$ $\gets$ ${\tt Classifier}(x)$
        \State recon\_dist\_x $\gets$ ${\tt Recon}(x,y_{pred})$
        \State pval $\gets$ $p$-${\tt value}(recon\_dist\_x,recon\_dists)$
        \If {pval $\leq$ $t$}
            \State Adversaries.${\tt insert}(x)$
        \EndIf
    \EndFor
    \State {\bf return} Adversaries 
\EndFunction
\Statex
\end{algorithmic}
  \vspace{-0.4cm}%
\end{algorithm}

% \todo[inline]{Describe the steps of the algorithm briefly here ...}
Algorithm~\ref{algo:adv_detect} first trains the CVAE network with clean training samples (Line~2) and formulates the reconstruction distances (Line~3). Then, for each of the test samples which may contain clean, randomly perturbed as well as adversarial examples, first the output predicted class is obtained using a target classifier network, followed by finding it's reconstructed image from CVAE, and finally by obtaining it's $p$-value to be used for thresholding (Lines~5-8). Images with $p$-value less than given threshold ($t$) are classified as adversaries (Lines~9-10).
% The histogram for distribution of p-values over different classes of adversaries and also for clean and randomly perturbed images for CIFAR-10 and MNIST datasets are shown in Figure~\ref{fig:p_cifar} and Figure~\ref{fig:p_mnist} respectively.  

\section{Experimental Results} \label{sec:experiment}
We experimented our proposed methodology over MNIST and CIFAR-10 datasets. All the experiments are performed in Google Colab GPU having $0.82$GHz frequency, $12$GB RAM and dual-core CPU having $2.3$GHz frequency, $12$GB RAM. An exploratory version of the code-base will be made public on github.
%also kept in the github repository for open access and execution
\subsection{Datasets and Models}
Two datasets are used for the experiments in this paper, namely MNIST~\cite{lecun2010mnist} and CIFAR-10~\cite{Krizhevsky09learningmultiple}. MNIST dataset consists of hand-written images of numbers from $0$ to $9$. It consists of $60,000$ training examples and $10,000$ test examples where each image is a $28 \times 28$ gray-scale image associated with a label from $1$ of the $10$ classes. CIFAR-10 is broadly used for comparison of image classification tasks. It also consists of $60,000$ image of which $50,000$ are used for training and the rest $10,000$ are used for testing. Each image is a $32 \times 32$ coloured image i.e. consisting of $3$ channels associated with a label indicating $1$ out of $10$ classes.

We use state-of-the-art deep neural network image classifier, ResNet18~\cite{he2015deep} as the target network for the experiments. We use the pre-trained model weights available from~\cite{Idelbayev18a} for both MNIST as well as CIFAR-10 datasets.

\subsection{Performance over Grey-box attacks}
If the attacker has the access only to the model parameters of the target classifier model and no information about the detector method or it's model parameters, then we call such attack setting as Grey-box. This is the most common attack setting used in previous works against which we evaluate the most common attacks with standard epsilon setting as used in other works for both the datasets. For MNIST, the value of $\epsilon$ is commonly used between 0.15-0.3 for FGSM attack and 0.1 for iterative attacks \cite{samangouei2018defensegan} \cite{gong2017adversarial} \cite{xu2017feature}. While for CIFAR10, the value of $\epsilon$ is most commonly chosen to be $\frac{8}{255}$ as in \cite{song2017pixeldefend} \cite{xu2017feature} \cite{fidel2020explainability}. For DeepFool \cite{moosavidezfooli2016deepfool} and Carlini Wagner (CW) \cite{carlini2017towards} attacks, the $\epsilon$ bound is not present. The standard parameters as used by default in \cite{li2020deeprobust} have been used for these 2 attacks. For $L_2$ attacks, the $\epsilon$ bound is chosen such that success of the attack is similar to their $L_\infty$ counterparts as the values used are very different in previous works.

\subsubsection{Reconstruction Error Distribution}
\begin{figure}[h]
\begin{center}
\begin{subfigure}{.4\textwidth}
    \centering
    \includegraphics[width=\textwidth]{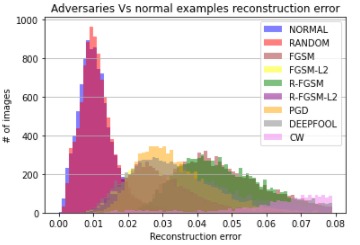}
    \caption{MNIST dataset}
\end{subfigure}
\begin{subfigure}{.4\textwidth}
    \centering
    \includegraphics[width=\textwidth]{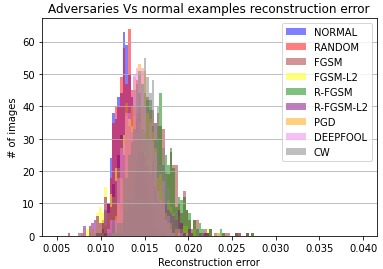}
    \caption{CIFAR-10 dataset}
\end{subfigure}
\caption{Reconstruction Distances for different Grey-box attacks}
\label{fig:recons_dist}
\end{center}
% \vspace{-0.5cm}
\end{figure}

The histogram distribution of reconstruction errors for MNIST and CIFAR-10 datasets for different attacks are given in Figure~\ref{fig:recons_dist}. For adversarial attacked examples, only examples which fool the network are included in the distribution for fair comparison. It may be noted that, the reconstruction errors for adversarial examples is higher than normal examples as expected. Also, reconstructions errors for randomly perturbed test samples are similar to those of normal examples but slightly larger as expected due to reconstruction error contributed from noise.

% \vspace{-0.3cm}

% \todo[inline]{Explain the distribution figures in more details here ...}
% Done

\subsubsection{$p$-value Distribution}
\begin{figure}[h] 
\begin{center}
    \begin{subfigure}{.4\textwidth}
        \centering
        \includegraphics[width=\textwidth]{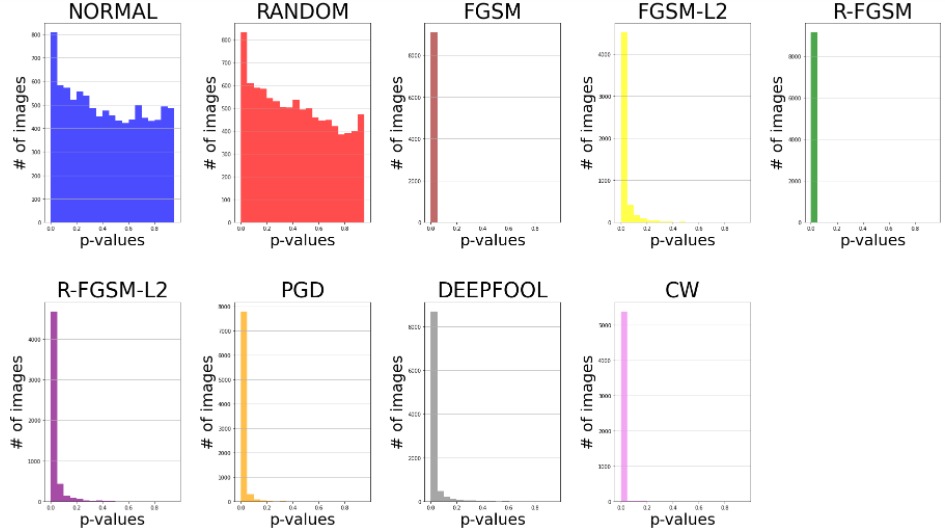}
        \caption{$p$-values from MNIST dataset}
        \label{fig:p_mnist}
    \end{subfigure}
    \begin{subfigure}{.4\textwidth}
        \centering
        \includegraphics[width=\textwidth]{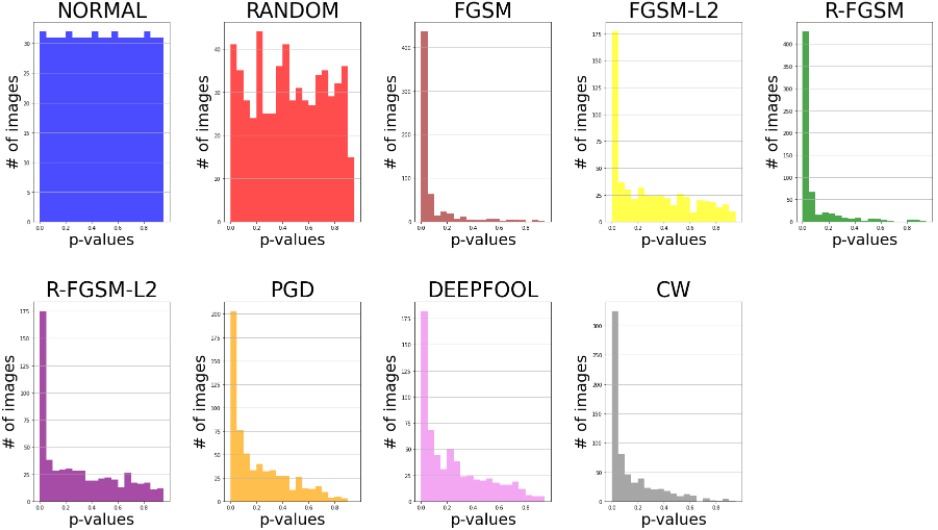}
        \caption{$p$-values from CIFAR-10 dataset}
        \label{fig:p_cifar}
    \end{subfigure}
    \caption{Generated $p$-values for different Grey-box attacks}
    \label{fig:p_val}
\end{center}
% \vspace{-0.4cm}
\end{figure}

From the reconstruction error values, the distribution histogram of p-values of test samples for MNIST and CIFAR-10 datasets are given in Figure~\ref{fig:p_val}. It may be noted that, in case of adversaries, most samples have $p$-value close to $0$ due to their high reconstruction error; whereas for the normal and randomly perturbed images, $p$-value is nearly uniformly distributed as expected. 

% \todo[inline]{Explain the distribution figures in more details here ...}
% Sir, I have already explained the details on how to construct above, shall I explain again here? It will consume a lot of space

\subsubsection{ROC Characteristics}
\begin{figure}[h] 
\begin{center}
    \begin{subfigure}{.38\textwidth}
        \centering
        \includegraphics[width=\textwidth]{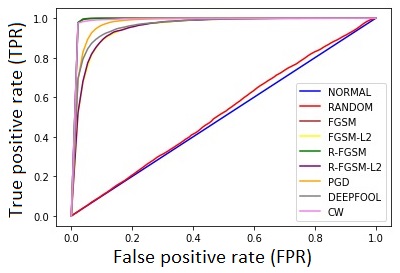}
        \caption{MNIST dataset}
        \label{fig:roc_mnist}
    \end{subfigure}
    \begin{subfigure}{.37\textwidth}
        \centering
        \includegraphics[width=\textwidth]{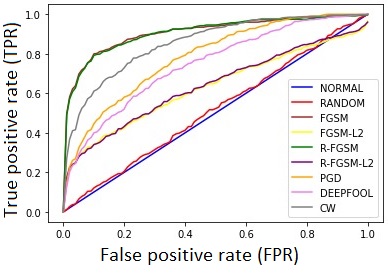}
        \caption{CIFAR-10 dataset}
        \label{fig:roc_cifar}
    \end{subfigure}
    \caption{ROC Curves for different Grey-box attacks}
    \label{fig:roc}
\end{center}
% \vspace{-1cm}
\end{figure}

Using the $p$-values, ROC curves can be plotted as shown in Figure \ref{fig:roc}. As can be observed from ROC curves, clean and randomly perturbed attacks can be very well classified from all adversarial attacks. The values of $\epsilon_{atk}$ were used such that the attack is able to fool the target detector for at-least $45\%$ samples. The percentage of samples on which the attack was successful for each attack is shown in Table~\ref{tab:stat}.

%%%%%%%%%%%%%% FIG HERE %%%%%%%%%%%%
% \vspace{-0.3cm}

% \todo[inline]{Explain the ROC figures in more details here ...}
% Done above 

\subsubsection{Statistical Results and Discussions}
The statistics for clean, randomly perturbed and adversarial attacked images for MNIST and CIFAR datasets are given in Table~\ref{tab:stat}. Error rate signifies the ratio of the number of examples which were misclassified by the target network. Last column (AUC) lists the area under the ROC curve. The area for adversaries is expected to be close to $1$; whereas for the normal and randomly perturbed images, it is expected to be around $0.5$.

% \vspace{-0.2cm}
\begin{table}[h]
{\sf \scriptsize
\begin{center}
\setlength\tabcolsep{1.4pt}
\begin{tabular}{|c|c|c|c|c|c|c|} 
 \hline
 {\bf Type} & \multicolumn{2}{c|}{\bf Error Rate (\%)} & \multicolumn{2}{c|}{\bf Parameters} & \multicolumn{2}{c|}{\bf AUC} \\
 \cline{2-3} \cline{4-5} \cline{6-7} 
  & {\bf MNIST} & {\bf CIFAR-10} & {\bf MNIST} & {\bf CIFAR-10} & {\bf MNIST} & {\bf CIFAR-10} \\
 \hline\hline
 NORMAL & 2.2 & 8.92 & - & - & 0.5 & 0.5\\ 
 \hline
 RANDOM & 2.3 & 9.41 & $\epsilon$=0.1 & $\epsilon$=$\frac{8}{255}$ & 0.52 & 0.514\\
 \hline
 FGSM & 90.8 & 40.02 & $\epsilon$=0.15 & $\epsilon$=$\frac{8}{255}$ & 0.99 & 0.91\\
 \hline
 FGSM-L2 & 53.3 & 34.20 & $\epsilon$=1.5 & $\epsilon=1$ & 0.95 & 0.63\\
 \hline
 R-FGSM & 91.3 & 41.29 & $\epsilon$=(0.05,0.1) & $\epsilon$=($\frac{4}{255}$,$\frac{8}{255}$) & 0.99 & 0.91\\
 \hline
 R-FGSM-L2 & 54.84 & 34.72 & $\epsilon$=(0.05,1.5) & $\epsilon$=($\frac{4}{255}$,1) & 0.95 & 0.64\\ 
 \hline
 PGD & 82.13 & 99.17 & $\epsilon$=0.1,$n$=12 & $\epsilon$=$\frac{8}{255}$,$n$=12 & 0.974 & 0.78\\
 &  &  & $\epsilon_{step}=0.02$ & $\epsilon_{step}$=$\frac{1}{255}$ &  & \\
 \hline
 CW & 100 & 100 & - & - & 0.98 & 0.86\\ 
 \hline
 DeepFool & 97.3 & 93.89 & - & - & 0.962 & 0.75\\
 \hline
\end{tabular}
\end{center}
}
\caption{Image Statistics for MNIST and CIFAR-10. AUC : Area Under the ROC Curve. Error Rate (\%) : Percentage of samples mis-classified or Successfully-attacked}
\label{tab:stat}
\end{table}
% \vspace{-0.3cm}

% \begin{table}[h]
% \begin{center}
% \begin{tabular}{|c||c|c|c|c|} 
%  \hline
%  {\bf Type} & {\bf Error Rate} & $\epsilon$ & {\bf Parameters} & {\bf AUC} \\
%  \hline\hline
%  Normal & 8.86\% & - & - & 0.5\\ 
%  \hline
%  RAND & 11.03\% & 20 & - & 0.53\\
%  \hline
%  FGSM & 50.17\% & 20 & - & 0.99\\
%  \hline
%  FGSM-L2 & 54.32\% & 30 & - & 0.97\\
%  \hline
%  R-FGSM & 50.52\% & 20 & $\epsilon$'=4 & 0.99\\
%  \hline
%  R-FGSM-L2 & 54.73\% & 30 & $\epsilon$'=4 & 0.97\\ 
%  \hline
%  BIM & 99.84\% & 20 & $\alpha$=4  & 0.85\\ 
%  \hline
%  BI-L2 & 99.66\% & 30 & $\alpha$=4 & 0.85\\ 
%  \hline
%  CW & 97.05\% & 20 & c=1 & 0.83\\ 
%  \hline
%  S-BIM & 94.32\% & 20 & $\alpha$=4, $\sigma$=0.6 & 0.5\\
%  \hline
% \end{tabular}
% \end{center}
% \caption{Image Statistics for CIFAR-10 Dataset}
% \label{tab:stat_cifar}
% \end{table}

It is worthy to note that, the obtained statistics are much comparable with the state-of-the-art results. Interestingly, some of the methods~\cite{song2017pixeldefend} explicitly report comparison results with randomly perturbed images and are ineffective in distinguishing adversaries from random noises, but most other methods do not report results with random noise added to the input image. Since other methods use varied experimental setting, attack models, different datasets as well as $\epsilon_{atk}$ values and network model, exact comparisons with other methods is not directly relevant primarily due to such varied experimental settings. However, the results are mostly similar to our results while our method is able to statistically differentiate from random noisy images.

\vspace{-0.2cm}
In addition to this, since our method does not use any adversarial examples for training, it is not prone to changes in value of $\epsilon$ or with change in attacks which network based methods face as they are explicitly trained with known values of $\epsilon$ and types of attacks. Moreover, among distribution and statistics based methods, to the best of our knowledge, utilization of the predicted class from target network has not been done before. Most of these methods either use the input image itself \cite{jha2018detecting} \cite{song2017pixeldefend} \cite{xu2017feature}, or the final logits layer \cite{feinman2017detecting} \cite{hendrycks2016early}, or some intermediate layer \cite{li2017adversarial} \cite{fidel2020explainability} from target architecture for inference, while we use the input image and the predicted class from target network to do the same.

% Unfortunately, there is no standard comparison metric available from previous works, hence a tabular comparison based on the claims from previous works is very difficult to produce since each method evaluates it's attack on different datasets, with different values of $\epsilon_{attack}$ for generating adversaries, different target classifier networks. The statistics for our work are presented in Table~\ref{}. 

% \todo[inline]{Explain the tables in more details here ...}
% Done already above

%%%%%%%%%%%%% TABLE HERE %%%%%%%%%%%%%%
\subsection{Performance over White-box attacks}
In this case, we evaluate the attacks if the attacker has the information of both the defense method as well as the target classifier network. \cite{metzen2017detecting} proposed a modified PGD method which uses the gradient of the loss function of the detector network assuming that it is differentiable along with the loss function of the target classifier network to generate the adversarial examples. If the attacker also has access to the model weights of the detector CVAE network, an attack can be devised to fool both the detector as well as the classifier network. The modified PGD can be expressed as follows :-
\begin{subequations}
\begin{flalign}
&X_{adv,0} = X,\\
&X_{adv,n+1} = {\tt Clip}_X^{\epsilon_{atk}}\Big{\{}X_{adv,n} + \nonumber\\
&\qquad \qquad \alpha .sign \big{(}\ (1-\sigma) . \Delta_X L_{cls}(X_{adv,n},y_{target}) + \nonumber\\
&\qquad \qquad  \sigma . \Delta_X L_{det}(X_{adv,n},y_{target})\ \big{)} \Big{\}}
\end{flalign}
\end{subequations}
Where $y_{target}$ is the target class and $L_{det}$ is the reconstruction distance from Equation \ref{eq:recon}. It is worthy to note that our proposed detector CVAE is differentiable only for the targeted attack setting. For the non-targeted attack, as the condition required for the CVAE is obtained from the target classifier output which is discrete, the differentiation operation is not valid. We set the target randomly as any class other than the true class for testing.  
\subsubsection{Effect of $\sigma$}
\begin{figure}[h] 
\begin{center}
    \begin{subfigure}{.4\textwidth}
        \centering
        \includegraphics[width=\textwidth]{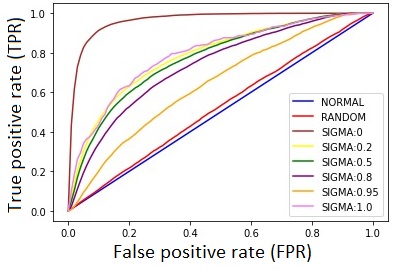}
        \caption{MNIST dataset}
        % \label{fig:roc_mnist}
    \end{subfigure}
    \begin{subfigure}{.4\textwidth}
        \centering
        % To be replaced
        \includegraphics[width=\textwidth]{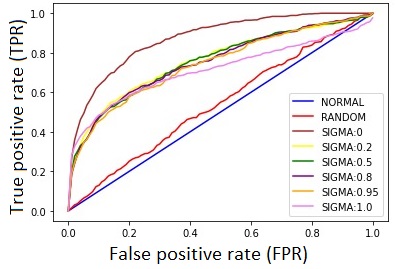}
        \caption{CIFAR-10 dataset}
        % \label{fig:roc_cifar}
    \end{subfigure}
    \caption{ROC Curves for different values of $\sigma$. More area under the curve implies better detectivity for that attack. With more $\sigma$ value, the attack, as the focus shifts to fooling the detector, it becomes difficult for the detector to detect.} 
    \label{fig:roc_sigma}
\end{center}
% \vspace{-1cm}
\end{figure}

\begin{figure}[h] 
\begin{center}
    \begin{subfigure}{.4\textwidth}
        \centering
        \includegraphics[width=\textwidth]{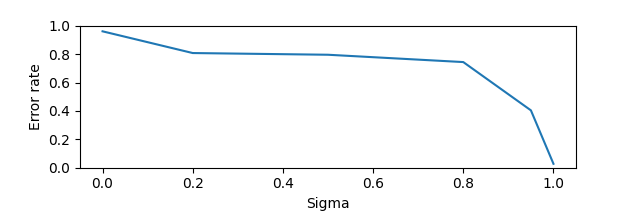}
        \caption{MNIST dataset}
        % \label{fig:roc_mnist}
    \end{subfigure}
    \begin{subfigure}{.4\textwidth}
        \centering
        % To be replaced
        \includegraphics[width=\textwidth]{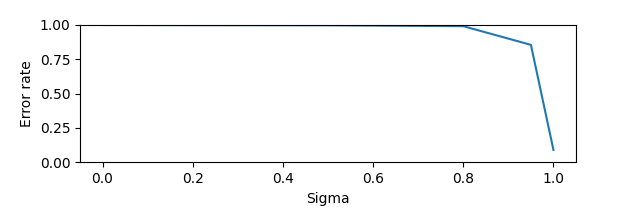}
        \caption{CIFAR-10 dataset}
        % \label{fig:roc_cifar}
    \end{subfigure}
    \caption{Success rate for different values of $\sigma$. More value of $\sigma$ means more focus on fooling the detector, hence success rate of fooling the detector decreases with increasing $\sigma$.}
    \label{fig:stats_sigma}
\end{center}
% \vspace{-1cm}
\end{figure}

To observe the effect of changing value of $\sigma$, we keep the value of $\epsilon$ fixed at 0.1. As can be observed in Figure \ref{fig:roc_sigma}, the increase in value of $\sigma$ implies larger weight on fooling the detector i.e. getting less reconstruction distance. Hence, as expected the attack becomes less successful with larger values of $\sigma$ \ref{fig:stats_sigma} and gets lesser AUC values \ref{fig:roc_sigma}, hence more effectively fooling the detector. For CIFAR-10 dataset, the detection model does get fooled for higher $c$-values but however the error rate is significantly low for those values, implying that only a few samples get attacks on setting such value.
\subsubsection{Effect of $\epsilon$}
\begin{figure}[h] 
\begin{center}
    \begin{subfigure}{.4\textwidth}
        \centering
        \includegraphics[width=\textwidth]{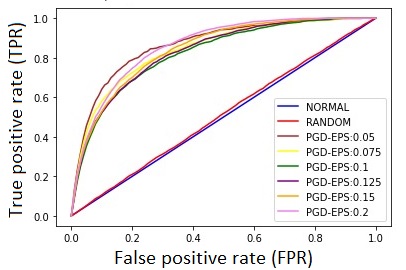}
        \caption{MNIST dataset}
        % \label{fig:roc_mnist}
    \end{subfigure}
    \begin{subfigure}{.4\textwidth}
        \centering
        \includegraphics[width=\textwidth]{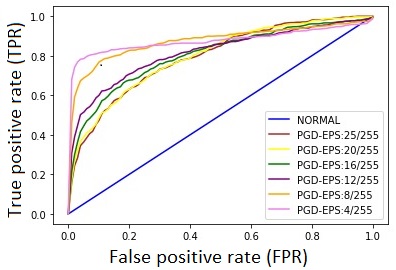}
        \caption{CIFAR-10 dataset}
        % \label{fig:roc_cifar}
    \end{subfigure}
    \caption{ROC Curves for different values of $\epsilon$. With more $\epsilon$ value, due to more space available for the attack, attack becomes less detectable on average.}
    \label{fig:roc_eps}
\end{center}
% \vspace{-1cm}
\end{figure}

\begin{figure}[h] 
\begin{center}
    \begin{subfigure}{.4\textwidth}
        \centering
        \includegraphics[width=\textwidth]{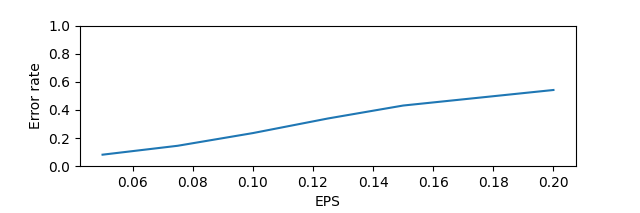}
        \caption{MNIST dataset}
        % \label{fig:roc_mnist}
    \end{subfigure}
    \begin{subfigure}{.4\textwidth}
        \centering
        \includegraphics[width=\textwidth]{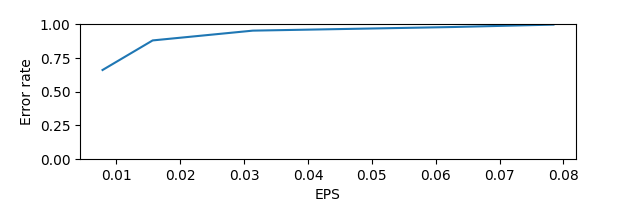}
        \caption{CIFAR-10 dataset}
        % \label{fig:roc_cifar}
    \end{subfigure}
    \caption{Success rate for different values of $\epsilon$. More value of $\epsilon$ means more space available for the attack, hence success rate increases}
    \label{fig:stats_eps}
\end{center}
% \vspace{-1cm}
\end{figure}

With changing values of $\epsilon$, there is more space available for the attack to act, hence the attack becomes more successful as more no of images are attacked as observed in Figure \ref{fig:stats_eps}. At the same time, the trend for AUC curves is shown in Figure \ref{fig:roc_eps}. The initial dip in the value is as expected as the detector tends to be fooled with larger $\epsilon$ bound. From both these trends, it can be noted that for robustly attacking both the detector and target classifier for significantly higher no of images require significantly larger attack to be made for both the datasets.

\section{Comparison with State-of-the-Art using Generative Networks}
Finally we compare our work with these 3 works \cite{meng2017magnet} \cite{hwang2019puvae} \cite{samangouei2018defensegan} proposed earlier which uses Generative networks for detection and purification of adversaries. We make our comparison on MNIST dataset which is used commonly in the 3 works (Table \ref{tab:stat2}). Our results are typically the best for all attacks or are off by short margin from the best. For the strongest attack, our performance is much better. This show how our method is more effective while not being confused with random perturbation as an adversary. 
\begin{table}[h]
{\sf \scriptsize
\begin{center}
\setlength\tabcolsep{2pt}
\begin{tabular}{|c|c|c|c|c|c|c|} 
 \hline
 {\bf Type} & \multicolumn{4}{c|}{\bf AUC} \\
 \cline{2-5}  
  & {\bf MagNet} & {\bf PuVAE} & {\bf DefenseGAN} & {\bf CVAE (Ours)} \\
 \hline\hline
 RANDOM & 0.61 & 0.72 & 0.52 & \textbf{0.52} \\
 \hline
 FGSM & 0.98 & 0.96 & 0.77 & \textbf{0.99} \\
 \hline
 FGSM-L2 & 0.84 & 0.60 & 0.60 & \textbf{0.95}\\
 \hline
 R-FGSM & \textbf{0.989} & 0.97 & 0.78 & 0.987\\
 \hline
 R-FGSM-L2 & 0.86 & 0.61 & 0.62 & \textbf{0.95}\\ 
 \hline
 PGD & \textbf{0.98} & 0.95 & 0.65 & 0.97\\
 \hline
 CW & 0.983 & 0.92 & 0.94 & \textbf{0.986}\\ 
 \hline
 DeepFool & 0.86 & 0.86 & 0.92 & \textbf{0.96} \\
 \hline
 \textbf{Strongest} & 0.84 & 0.60 & 0.60 & \textbf{0.95}\\ 
 \hline
\end{tabular}
\end{center}
}
\caption{Comparison in ROC AUC statistics with other methods. More AUC implies more detectablity. 0.5 value of AUC implies no detection. For RANDOM, value close to 0.5 is better while for adversaries, higher value is better.}
\label{tab:stat2}
\end{table}

\subsection{Use of simple AutoEncoder (AE)}
\begin{figure}[h] 
\begin{center}
    \includegraphics[width=.3\textwidth]{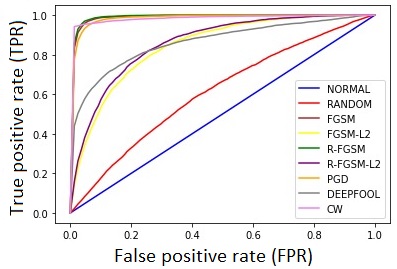}
    \includegraphics[width=.3\textwidth]{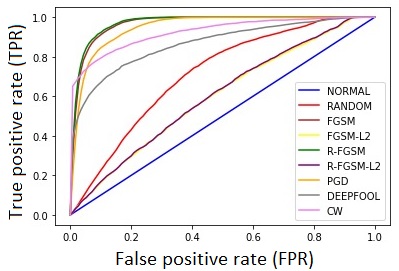}
    \includegraphics[width=.3\textwidth]{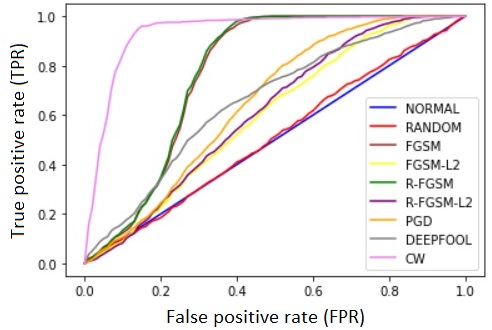}
    \caption{ROC curve of different adversaries for MagNet (left), PuVAE (centre), DefenseGAN (right)}
    \label{fig:ae}
\end{center}
% \vspace{-1cm}
\end{figure}

MagNet \cite{meng2017magnet} uses AutoEncoder (AE) for detecting adversaries. We compare the results with our proposed CVAE architecture on the same experiment setting and present the comparison in AUC values of the ROC curve observed for the 2 cases. Although the paper's claim is based on both detection as well as purification (if not detected) of the adversaries. MagNet uses their detection framework for detecting larger adversarial perturbations which cannot be purified. For smaller perturbations, MagNet proposes to purify the adversaries by a different AutoEncoder model. We make the relevant comparison only for the detection part with our proposed method. Using the same architecture as proposed, our results are better for the strongest attack while not getting confused with random perturbations of similar magnitude. ROC curves obtained for different adversaries for MagNet are given in Figure \ref{fig:ae}      
\subsection{Use of Variational AutoEncoder (VAE)}
PuVAE \cite{hwang2019puvae} uses Variational AutoEncoder (VAE) for purifying adversaries. We compare the results with our proposed CVAE architecture on the same experiment setting. PuVAE however, does not propose using VAE for detection of adversaries but in case if their model is to be used for detection, it would be based on the reconstruction distance. So, we make the comparison with our proposed CVAE architecture. ROC curves for different adversaries are given in Figure \ref{fig:ae}

\subsection{Use of Generative Adversarial Network (GAN)}
Defense-GAN \cite{samangouei2018defensegan} uses Generative Adversarial Network (GAN) for detecting adversaries. We used $L=100$ and $R=10$ for getting the results as per our experiment setting. We compare the results with our proposed CVAE architecture on the same experiment setting and present the comparison in AUC values of the ROC curve observed for the 2 cases. Although the paper's main claim is about purification of the adversaries, we make the relevant comparison for the detection part with our proposed method. We used the same architecture as mentioned in \cite{samangouei2018defensegan} and got comparable results as per their claim for MNIST dataset on FGSM adversaries. As this method took a lot of time to run, we randomly chose 1000 samples out of 10000 test samples for evaluation due to time constraint. The detection performance for other attacks is considerably low. Also, Defense-GAN is quite slow as it needs to solve an optimization problem for each image to get its corresponding reconstructed image. Average computation time required by Defense-GAN is $2.8s$ per image while our method takes $0.17s$ per image with a batch size of $16$. Hence, our method is roughly 16 times faster than Defense-GAN. Refer to Figure \ref{fig:ae} for the ROC curves for Defense-GAN.

\chapter{Use of Variational AutoEncoder (VAE) for Purification}

\section{Introduction}
After detecting adversaries next net logical thing to do is to purify the images before passing to the target classifier. The idea of purification is to neutralise or reduce the threat of any adversary attack if made on the image. There are various existing works doing this. MagNet .... PuVAE..... PixelDefend..... DefenseGAN....... We however propose use of Variational AutoEncoder similar to MagNet but however with a modified form similar to DefenseGAN and PixelDefend.  

\section{Training of Variational AutoEncoder (VAE)}
The proposed Variational AutoEncoder is trained similar to MagNet on only clean images. The network architecture consists of an $Encoder$ and a $Decoder$. $Encoder$ outputs mean, $\mu$ and variance, $\sigma$ of size, latent dimension, $z$. A value is sampled from the obtained value of $z$ and is fed to the decoder to output a reconstructed image $X_{recon}$ os the same size as $X$. The training loss, $L$ is described as follows :-

\begin{eqnarray} \label{kuch_bhi}
    L(X) = &BCE[X,Decoder(z)] - \\
    & \frac{1}{2}[Encoder_{\sigma}^2(X)+Encoder_{\mu}^2(X)-1-log(Encoder_{\sigma}^2(X))]
\end{eqnarray}

\section{Purification of Input Images}
For purification of an input image, unlike MagNet \cite{meng2017magnet} instead of directly taking the reconstructed image, $X_{recon}$ as the purified image to feed to the classifier, we propose solving the following objective to achieve the same where $X_{purified}$ is the resultant purified image :-
\begin{equation}
    X_{purified} = \min\limits_{\epsilon} c.\left\Vert\epsilon\right\Vert_2^2 + \left\Vert X+\epsilon-recon(X+\epsilon)\right\Vert_2^2
\end{equation} \label{eqn_pur}
where $c$ is a constant parameter chosen accordingly. We use ADAM optimizer with $n$ iterations for each image. 

\section{Results on MNIST and Comparison With Other Works on MNIST dataset}
We present the comparison in results for MNIST dataset \cite{lecun2010mnist} with the 2 methods discussed earlier MagNet and DefenseGAN. Results for adversarial training are generated in the same way as described in MagNet \cite{meng2017magnet}. All results are with $\epsilon=0.1$.

\begin{table}[h]
{\sf \scriptsize
\begin{center}
\begin{tabular}{|c|c|c|c|c|c|}
    \hline
  {\bf Attack}  & {\bf No defense}  & {\bf Adversarial training} & {\bf DefenseGAN} & {\bf MagNet} & {\bf Ours}   \\
  \hline
    Clean  &  0.974 & 0.822 & 0.909 & 0.95 & \bf{0.97} \\ 
  \hline
    FGSM  &  0.269 & 0.651 & 0.875 & 0.63 & \bf{0.887} \\ 
  \hline
    R-FGSM  &  0.269 & 0.651 & 0.875 & 0.63 & \bf{0.887} \\ 
  \hline
    PGD  &  0.1294 & 0.354 & 0.83 & 0.65 & \bf{0.886} \\ 
  \hline
    CW  & 0.1508 & 0.28 & 0.656 & 0.85 & \bf{0.907} \\ 
\hline
\end{tabular}
\end{center}
}
\caption{Comparison in results. Values reported are classifier success rate (in fraction out of 1)}
\label{tab:comp_res}
\end{table}

\section{Variations for CIFAR-10 Dataset}

MNIST dataset is a simple dataset with easily identifiable number shapes which can be easily and robustly learned by a neural network. It is therefore very easy to correct a perturbed image with a simple back-propagation method with fixed no of iterations. For CIFAR-10 dataset however, due to the inherent complexity in classifying objects, it is very easy to attack and thus simple defenses with fixed no of iterations is not enough to give a reasonably good purification result on attacks.

\subsection{Fixed no of iterations}

First, we evaluate the results with fixed no of iterations. The results are listed on Table \ref{tab:comp_cifar}. As can be observed with more no of iterations clean accuracy gets reduced and adversarial accuracy improves. The drawback with such method is that it is not essentially using any discrimination for the update rule for adversarial and non adversarial examples. As a result of this, reconstruction errors of both adversarial and clean images (see Figure \ref{fig:recons_errors_A}) are reduced leading to disappearance of important features. Formally, the purified example, $X_{pur}$ can be expressed as follows where $X_{adv}$ is the adversarial image, $n$ is the fixed no of iterations, $\alpha$ is the constant learning rate and $purifier$ is the Purifier autoencoder function. 

\begin{equation}
\begin{split}
    &X_{pur,0} = X_{adv} \\
    &X_{pur,i+1} = X_{pur,i} - \alpha \frac{\partial L(X_{pur,i},purifier(X_{pur,i}))}{\partial X_{pur,i}} \text{ for } i \in \{1,2...,n\}\\
    &\text{Where } L(X,Y) = (X-Y)^2 \\
\end{split} \label{Eqn:fix_iter_basic}
\end{equation}

\subsection{Fixed no of iterations with using ADAM optimizer for update}

Major drawback with using constant learning rate is that both clean and adversarial examples are updated for fixed no of pixels in the image space meaning the distance between input and purified image is same irrespective if the input is adversarial or not. To counter it, we use ADAM optimizer to achieve the minima quickly. with same no of iterations. In this case, the update amount varies for adversarial and clean image as clean image has less reconstruction error (see Figure \ref{fig:recons_errors_B}). Formally, the update is defined as follows where $X_{adv}$ is the adversarial image, $X_{pur}$ is the purified image, $n$ is the no of iterations, $\alpha$ and $\beta$ are parameters for ADAM optimizer.

\begin{equation}
\begin{split}
    &X_{pur,0} = X_{adv} \\
    &X_{pur,i+1} = X_{pur,i} - \alpha w_i\\
    &\text{Where } w_i = \beta w_{i-1} + (1-\beta) \frac{\partial L(X_{pur,i},purifier(X_{pur,i}))}{\partial X_{pur,i}} \text{ for } i \in \{1,2...,n\}\\
    &\text{Where } L(X,Y) = (X-Y)^2 \\
\end{split} \label{Eqn:fix_iter_adam}
\end{equation}

\subsection{Variable learning rate based on the current reconstruction error}

Usually for adversarial images, the initial reconstruction error is high as compared to clean image. Based on the following, an advantage can be taken to purify mostly only the adversarial images and not the clean images. No of iterations can be varied based on the reconstruction error of the input sample. First, the mean value, $\mu$ and variance, $\sigma$ of the distribution of reconstruction errors for the clean validation samples are determined. Then, the probability of falling within the equivalent gaussian distribution of reconstruction distances of clean validation samples is determined. The learning rate is varied accordingly linear to this probability. Formally, update rule can be defined as follows where $X_{adv}$ is the adversarial image, $X_{pur}$ is the purified image, $n$ is the no of iterations, $\alpha$ and $\beta$ are parameters for ADAM optimizer.

\begin{equation}
\begin{split}
    &X_{pur,0} = X_{adv} \\
    &X_{pur,i+1} = X_{pur,i} - \alpha_i w_i \\
    &\text{Where } \alpha_i = \alpha (1 - \exp^{-(\frac{X_{pur,i}-\mu}{\sigma})^2}) \\
    &\text{And } w_i = \beta w_{i-1} + (1-\beta) \frac{\partial L(X_{pur,i},purifier(X_{pur,i}))}{\partial X_{pur,i}} \text{ for } i \in \{1,2...,n\}\\
    &\text{Where } L(X,Y) = (X-Y)^2 \\
\end{split} \label{Eqn:fix_iter_adam_variable_lr}
\end{equation}

\subsection{Set target distribution for reconstruction error}

Major drawback with earlier variations is that even though they discriminate in purifying already clean and adversarial images, they still try to purify clean images by bringing down the reconstruction error. One drawback to this is on doing so, the important features of the image are lost as the purification model sees them as perturbations to the images giving rise to the reconstruction error. Back-propagating through them smoothens them leading to less reconstruction error but confuses the classifier leading to wrong predictions. To avoid this, we first find the mean, $\mu$ and variance, $\sigma$ of the clean image validation set. The objective for the update rule is kept to increase the probability mass function value with respect to the target distribution. Formally, the update rule is defined as follows :-

\begin{equation}
\begin{split}
    &X_{pur,0} = X_{adv} \\
    &X_{pur,i+1} = X_{pur,i} - \alpha w_i\\
    &\text{Where } w_i = \beta w_{i-1} + (1-\beta) \frac{\partial L(X_{pur,i},purifier(X_{pur,i}))}{\partial X_{pur,i}} \text{ for } i \in \{1,2...,n\}\\
    &\text{Where } L(X,Y) = \frac{|dist(X,Y)-mu|}{\sigma} \\
    &\text{Where } dist(X,Y) = (X-Y)^2
\end{split} \label{Eqn:fix_iter_target}
\end{equation}

\subsection{Set target distribution for reconstruction error with modified update rule}

\begin{figure}[h] 
    \begin{center}
    \begin{subfigure}{.4\textwidth}
        \centering
        \includegraphics[width=\textwidth]{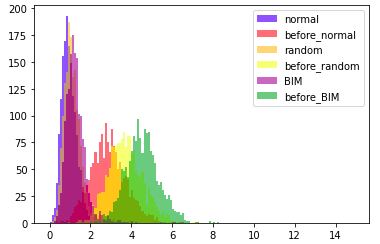}
        \caption{Fixed no of iterations}
        \label{fig:recons_errors_A}
    \end{subfigure}
    \begin{subfigure}{.4\textwidth}
        \centering
        \includegraphics[width=\textwidth]{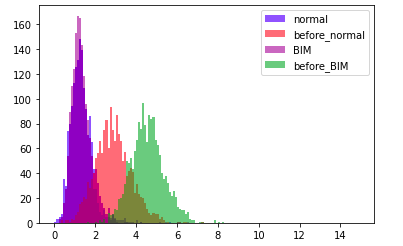}
        \caption{Fixed no of iterations with using ADAM optimizer for update}
        \label{fig:recons_errors_B}
    \end{subfigure}
    \begin{subfigure}{.4\textwidth}
        \centering
        \includegraphics[width=\textwidth]{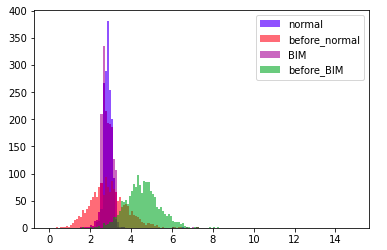}
        \caption{Set target distribution for reconstruction error}
        \label{fig:recons_errors_C}
    \end{subfigure}
    \begin{subfigure}{.4\textwidth}
        \centering
        \includegraphics[width=\textwidth]{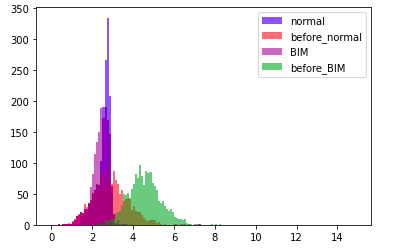}
        \caption{Set target distribution for reconstruction error with modified update rule}
        \label{fig:recons_errors_D}
    \end{subfigure}
    \caption{Reconstruction errors change for different variations}
    \label{fig:recon_errors_mnist}
    \end{center}
\end{figure}

Drawback of the above method is that it tries to increase the reconstruction error (see Figure \ref{fig:recons_errors_C}) of the samples with less reconstruction error belonging to clean image set. Due to this, classification model gives less accuracy. To avoid this, we change the update rule as follows which ultimately leads to no change for clean images with reconstruction error less than $\mu$ (see Figure \ref{fig:recons_errors_C}). 

\begin{equation}
\begin{split}
    &X_{pur,0} = X_{adv} \\
    &X_{pur,i+1} = X_{pur,i} - \alpha w_i\\
    &\text{Where } w_i = \beta w_{i-1} + (1-\beta) \frac{\partial L(X_{pur,i},purifier(X_{pur,i}))}{\partial X_{pur,i}} \text{ for } i \in \{1,2...,n\}\\
    &\text{Where } L(X,Y) = \frac{max(dist(X,Y)-mu,0)}{\sigma} \\
    &\text{Where } dist(X,Y) = (X-Y)^2
\end{split} \label{Eqn:fix_iter_target_modified}
\end{equation}

\subsection{Add random noise at each update step}

Just adding random noise has been observed to improve the classification accuracy for adversarial examples as it changes the overall perturbation to near random breaking the targeted perturbation created by the adversarial attack to the corresponding clean sample. This technique can be used in conjunction to our method where random perturbation is added at each update step. This leads to slightly better results as observed in Table \ref{tab:comp_cifar}. Formally the update rule is defined as follows where $\gamma$ is the amount of noise to be added at each update step.

\begin{equation}
\begin{split}
    &X_{pur,0} = X_{adv} \\
    &X_{pur,i+1} = X_{pur,i} - \alpha w_i\\
    &\text{Where } w_i = \beta w_{i-1} + (1-\beta) \frac{\partial L(X_{pur,i},purifier(X_{noise,pur,i}))}{\partial X_{noise,pur,i}} \text{ for } i \in \{1,2...,n\}\\
    &\text{Where } X_{noise,pur,i} = X_{pur,i} + \gamma r_X, r_X ~ \mathcal{N}(0,I_X) \\
    &\text{Where } L(X,Y) = \frac{max(dist(X,Y)-mu,0)}{\sigma} \\
    &\text{Where } dist(X,Y) = (X-Y)^2
\end{split} \label{Eqn:fix_iter_target_random}
\end{equation}

\subsection{Add random transformation at each update step}

Adding random rotate and resize transformations to the input image has also been observed to improve the classification accuracy for adversarial examples \cite{8954476}. This technique can be used in conjunction to our method where random transformation is added at each update step. This leads to slightly better results as observed in Table \ref{tab:comp_cifar}. Formally the update rule is defined as follows where $t$ is the transformation function taking the resize factor $f$ and rotation $\theta$ as input.

\begin{equation}
\begin{split}
    &X_{pur,0} = X_{adv} \\
    &X_{pur,i+1} = X_{pur,i} - \alpha w_i\\
    &\text{Where } w_i = \beta w_{i-1} + (1-\beta) \frac{\partial L(X_{pur,i},purifier(X_{trans,pur,i}))}{\partial X_{trans,pur,i}} \text{ for } i \in \{1,2...,n\}\\
    &\text{Where } X_{trans,pur,i} = t(X_{pur,i},f,\theta) \\
    &\text{Where } L(X,Y) = \frac{max(dist(X,Y)-mu,0)}{\sigma} \\
    &\text{Where } dist(X,Y) = (X-Y)^2
\end{split} \label{Eqn:fix_iter_target_trans}
\end{equation}

\begin{table}[h]
{\sf \scriptsize
\begin{center}
\begin{tabular}{|m{1.1cm}|m{0.9cm}|m{1.5cm}|m{1cm}|m{0.75cm}|m{0.75cm}|m{0.75cm}|m{0.5cm}|m{0.5cm}|m{0.6cm}|m{0.6cm}|m{0.5cm}|m{0.5cm}|m{1cm}|}
    \hline
  {\bf Attack}  & {\bf No defense}  & {\bf Adversarial training} & {\bf MagNet} & {\bf A ($n$=12)} & {\bf A ($n$=15)} & {\bf A ($n$=18)} & {\bf B} & {\bf C} & {\bf D} & {\bf E} & {\bf F} & {\bf G} & {\bf H (Ours)}   \\
  \hline
    Clean  &  0.954 & 0.871 & 0.790 & 0.879 & 0.853 & 0.843 & 0.872 & 0.892 & \bf{0.924} & 0.907 & 0.908 & 0.858 & 0.905\\
  \hline
    Random  &  0.86 & - & 0.793 & 0.889 & 0.875 & 0.867 & 0.883 & 0.874 & \bf{0.90} & 0.873 & 0.893 & 0.846 & 0.889\\
  \hline
    FGSM  &  0.533 & 0.650 & 0.661 & 0.73 & 0.76 & 0.774 & 0.778 & 0.769 & 0.783 & \bf{0.834} & 0.818 & 0.828 & 0.825\\ 
  \hline
    R-FGSM  &  0.528 & 0.640 & 0.655 & 0.73 & 0.747 & 0.764 & 0.772 & 0.762 & 0.767 & \bf{0.817} & 0.811 & 0.814 & 0.815\\ 
  \hline
    BIM  &  0.002 & 0.483 & 0.367 & 0.632 & 0.694 & 0.702 & 0.706 & 0.692 & 0.684 & 0.702 & 0.729 & 0.711 & \bf{0.735}\\
\hline
\end{tabular}
\end{center}
}
\caption{Comparison of results for different variations on Cifar-10 dataset. A : Fixed no of iterations, B : Fixed no of iterations with using ADAM optimizer for update, C : Variable learning rate based on the current reconstruction error, D : Set target distribution for reconstruction error, E : Set target distribution for reconstruction error with modified update rule, F : Add random noise at each update step, G : Add random transformation at each update step, H : Add random noise + transformation at each update step}
\label{tab:comp_cifar}
\end{table}

\section{Variations for ImageNet Dataset}

ImageNet dataset is the high resolution dataset with cropped RGB images of size 256X256. For our experiments we use pretrained weights on ImageNet train corpus available from \cite{5206848}. For test set we use the 1000 set of images available for ILSVRC-12 challenge \cite{Russakovsky2015ImageNetLS} which is commonly used by many works \cite{xu2017feature} for evaluating adversarial attack and defense methods. This test set consists of 1000 images each belonging to a different class. The pre-trained weights of ResNet-18 classifier are also provided by ILSVRC-12 challenge \cite{Russakovsky2015ImageNetLS}. The 1000 images are chosen such that the classifier gives correct class for all of them. 

\subsection{Attack method}

The attack method for ImageNet dataset is different as the classification accuracy for ImageNet dataset comprising of 1000 classes is defined by the top-1 accuracy which means a prediction is correct if any of the top 1\% of total classes or 10 class predictions in this case are correct. Hence, the iterative attack (which is used as baseline by all methods) is defined mathematically as follows where $J(.)$ is the cross entropy loss, $classifier(.)$ is the target classifier neural network, $\alpha$ is the step size and $\pi_{X,\epsilon}(.)$ function restricts value within $[X-\epsilon,X+\epsilon]$.

\begin{equation}
\begin{split}
    &X_{attacked,0} = X_{cln} \\
    &X_{attacked,i+1} = \pi_\epsilon[X_{attacked,i} + \alpha \frac{\partial J(classifier(X_{attacked,i}),y)}{\partial X_{attacked,i}}] \text{ for } i \in \{1,2...,n\}\\
\end{split} \label{Eqn:imagenet_attack}
\end{equation}

\subsection{Noise remover network}

We use autoencoder with skip connections and blind spot architecture for noise removal \cite{Laine2019HighQualitySD}. The network is trained as self-supervised task in N2V manner with the task of outputting a clean image given a noisy image. For additional details regarding the training settings for unknown noise form and magnitude, readers are referred to \cite{Laine2019HighQualitySD}. 

\subsection{No of iterations}

Effect of varying the no of iterations of update for purifying adversarial examples can be seen in Table \ref{tab:comp_imagenet}. As can be observed, increasing no of iterations lead to better classification accuracy for purified adversaries but leads to drop in clean accuracy. Results for adversaries created with different no of iterations are also reported. The value of $\epsilon=\frac{8}{255}$ and update step, $\alpha=\frac{1}{255}$ are chosen as standard values for comparison with \cite{xu2017feature}.

\subsection{Purification variations}

We further explore different variations to the base version with fixed no of iterations and constant update rate $\alpha$. 

\subsubsection{Random noise at each step}

Similar to Cifar-10 dataset, adding random noise to input leads to increase in classification accuracy. Hence, we conjunct this by adding random noise at each update step. As evident from Table \ref{tab:comp_imagenet}, there is a slight surge in classification of purified adversaries with this addition. Mathematically, update rule is defined as follows where $\gamma$ is the amount of noise to be added at each update step.

\begin{equation}
\begin{split}
    &X_{pur,0} = X_{adv} \\
    &X_{pur,i+1} = X_{pur,i} - \alpha \frac{\partial L(X_{pur,i},purifier(X_{noise,pur,i}))}{\partial X_{noise,pur,i}} \text{ for } i \in \{1,2...,n\}\\
    &\text{Where } X_{noise,pur,i} = X_{pur,i} + \gamma r_X, r_X ~ \mathcal{N}(0,I_X) \\
    &\text{Where } L(X,Y) = (X-Y)^2
\end{split} \label{Eqn:imagenet_random}
\end{equation}

\subsubsection{Random transformation at each step}

Similar to Cifar-10 dataset, we add random rotate and resize transformation before each update step. This too results in a subsequent increase in classification accuracy of purified adversaries as observed in Table \ref{tab:comp_imagenet}. The reason for the rise in accuracy is that the transformation leads to change in the locality of the features where targeted attack was made. The classification network however has been trained to act robustly if the image is resized or rotated, hence the classified class for clean images doesn't change, while for attacked images due to relocation of targeted perturbations, the resultant perturbation tends towards non-targeted random noise leading to increased accuracy. Mathematically, update rule can be expressed as follows where $t$ is the transformation function taking the resize factor $f$ and rotation $\theta$ as input.

\begin{equation}
\begin{split}
    &X_{pur,0} = X_{adv} \\
    &X_{pur,i+1} = X_{pur,i} - \alpha \frac{\partial L(X_{pur,i},purifier(X_{trans,pur,i}))}{\partial X_{trans,pur,i}} \text{ for } i \in \{1,2...,n\}\\
    &\text{Where } X_{trans,pur,i} = t(X_{pur,i},f,\theta) \\
    &\text{Where } L(X,Y) = (X-Y)^2
\end{split} \label{Eqn:imagenet_trans}
\end{equation}

\begin{table}[h]
{\sf \scriptsize
\begin{center}
\begin{tabular}{|c|c|c|c|c|c|c|}
    \hline
  {\bf Attack}  & {\bf No defense}  & {\bf Adversarial training} & {\bf MagNet} & {\bf A} & {\bf B} & {\bf C}   \\
  \hline
    Clean  &  1.0 & 0.912 & \bf{0.916} & 0.902 & 0.895 & 0.892 \\ 
  \hline
    Random  &  0.969 & 0.901 & \bf{0.911} & 0.907 & 0.887 & 0.884 \\ 
  \hline
    BIM ($\epsilon=\frac{10}{255}$)  &  0.361 & 0.441 & 0.539 & 0.670 & 0.705 & \bf{0.726} \\
  \hline
    BIM ($\epsilon=\frac{25}{255}$)  &  0.002 & 0.254 & 0.312 & 0.580 & 0.665 & \bf{0.681}\\
\hline
\end{tabular}
\end{center}
}
\caption{Comparison of results for different variations on Imagenet dataset. A : Ours, B : Ours (With random noise), C : Ours (With random transformations)}
\label{tab:comp_imagenet}
\end{table}

\section{Possible Counter Attacks}

Based on the attack method, adaptive attacks can be developed to counter the attack. We study the 2 types of attacks in detail. For a detailed review on how to systematically determine an adaptive attack based on the category of defense, readers are referred to \cite{Laine2019HighQualitySD}.

\subsection{Counter attack A}

The first adaptive attack possible is designed by approximating the transformation function i.e. the function of the output image obtained by modifying the input image through the update rule by the differentiable function obtained by autoencoder end-to-end output given the input. Intuitively this can be thought of as the near output obtained by backpropagating through the reconstruction error loss between the input and end-to-end purified output after passing through autoencoder purifier network. We observe (see Table \ref{tab:comp_counter}) that on applying this counter attack, the accuracy for direct purifier output method drops drastically while ours method gives better robust results against this attack. Mathematically, we can express the attacked image, $X_{attacked}$ as follows where $X_{cln}$ is the original clean image, $n$ is the no of iterations, $classifier(.)$ is the target classifier network, $J(.)$ is the cross entropy loss, $purifier(.)$ is the purifier autoencoder, $\alpha$ is the update rate and $\pi_{X,\epsilon}(.)$ function restricts value within $[X-\epsilon,X+\epsilon]$.

\begin{equation}
\begin{split}
    &X_{attacked,0} = X_{cln} \\
    &X_{attacked,i+1} = \pi_{X_{cln},\epsilon}[X_{attacked,i} + \alpha sign(\frac{\partial J(classifier(X_{attacked,i}),y)}{\partial X_{attacked,i}})] \\
    &\text{ for } i \in \{1,2...,n\}
\end{split} \label{Eqn:counter_attack_A}
\end{equation}

\subsection{Counter attack B}

The second adaptive attack possible here is by modifying the loss function of the BIM attack by including new weighted term getting less reconstruction error from the purifier This way we attack both the classifier as well as purifier. The purifier method relies on reconstruction error as a measure to update the input to get less reconstruction error similar to clean images but if the attack is made with this consideration to fool the purifier method by giving similar reconstruction error to clean image while also fooling the classifier, the attack is successful as it bypasses the purifier method. For this attack, the attacked image seems to be attacked more at the edges as modifying those do not change the reconstruction error for purifier much. As observed from Table \ref{tab:comp_counter}, the adaptive counter attack is successful to an extent but needs a larger value of $\epsilon$ to make the attack have a considerably successful attack rate. Mathematically this attack is defined as follows where Mathematically, we can express the attacked image, $X_{attacked}$ as follows where $X_{cln}$ is the original clean image, $n$ is the no of iterations, $classifier(.)$ is the target classifier network, $J(.)$ is the cross entropy loss, $purifier(.)$ is the purifier autoencoder, $\alpha$ is the update rate, $\beta$ is the weighing factor for the reconstruction error in the combined loss function, and $\pi_{X,\epsilon}(.)$ function restricts value within $[X-\epsilon,X+\epsilon]$.

\begin{equation}
\begin{split}
    &X_{attacked,0} = X_{cln} \\
    &X_{attacked,i+1} = \pi_{X_{cln},\epsilon}[X_{attacked,i} + \alpha sign(\frac{\partial J(X_{attacked,i},y) + \beta (X_{attacked,i} - purifier(X_{attacked,i}))^2}{\partial X_{attacked,i}})] \\
    &\text{ for } i \in \{1,2...,n\}
\end{split} \label{Eqn:counter_attack_B}
\end{equation}

\begin{table}[h]
{\sf \scriptsize
\begin{center}
\begin{tabular}{|c|c|c|c|c|c|}
    \hline
  {\bf Attack}  & {\bf No defense}  & {\bf Adversarial training} & {\bf MagNet} & {\bf Ours}   \\
  \hline
    Counter attack A  &  0.004 & 0.122 & 0.199 & \bf{0.680} \\ 
  \hline
    Counter attack B  &  0.009 & 0.151 & 0.275 & \bf{0.507} \\ 
\hline
\end{tabular}
\end{center}
}
\caption{Comparison of results for counter attacks}
\label{tab:comp_counter}
\end{table}

\chapter {Future Work}

Further, I plan to complete experimentation with the proposed VAE method for purrifying adversaries and try t beat the benchmark for results on other datasets like CIFAR-10, CIFAR-100, especially high dimensional ImageNet etc. Also, I would extend my proposed CVAE for detection to test on ImageNet. ImageNet has been especially challenging to get good results because of the high dimensional images, the attack perturbation is quite less.

% Although, the current state of the art detection methods have shown considerable robustness in detecting adversaries of various types on many known datasets, however there is a lot of limitation in terms of performance. The models currently are not able to generalize on all types of attacks. They work well for simple gradient based adversaries like FGSM, BIM etc but fail to recognise complex adversaries. Also, a lot of methods are currently focused on detecting white box adversaries, detection of black box adversaries has not been explored much yet. Most of the method do work well on some specific types of adversaries or on some specific dataset, but not able to generalize well along datasets and types of adversaries. With the curren detection framework. Most of the methods intentionally or not rely on gradient masking as the criteria. Athalye et al. [55] suggested several tips to identify gradient masking based detection: 1) if single step attacks perform better than iterative attacks, 2) if blackbox attacks perform better than whitebox attacks, 3) if unbounded perturbations do not reach 100\% attack success rate, and 4) if random sampling successfully finds adversarial examples while gradient based attacks do not. Following observations suggest that there is still a whole lot of scope on the future work along the line of adversary detection

Apart from this, the recent work by Pang et al. \cite{pang2018robust} suggests that training the network using RCE is much more robust than the currently use cross entropy loss. This opens a completely new direction for research. There can be a separate RCE trained network for the sake of robustness as the detector. There can be a corresponding research along these lines by training a generator network from logits with this new loss function as it is relatively robust against adversaries.

Also, the use of LSTM for detecting abnormal paths taken by adversaries along the distance vector spaces seems to be an interesting research direction. The idea can be further explored by using maybe transformers instead of LSTM to model the pattern in the path taken. Also, instead of taking mean vectors from the class distribution, clustering methods like K means clustering can be used to get mean vectors for the clusters. Also, Instead of using distant spaces, PCA or similar method can be used to map to a low dimensional representation from each layer.

Another interesting approach would be using representation learning for bayesian inference. Currently for the distribution based methods, logits vector, deep representations or input have been used for bayesian inference. What would be interesting is the use of unsupervised representation learning to get mapping of the input to a distribution which can be used for figuring out whether a given sample falls within the normal sample distribution or the adversarial examples.

\addcontentsline{toc}{chapter}{Future Work}
\printbibliography

@misc{amodei2016concrete,
      title={Concrete Problems in AI Safety}, 
      author={Dario Amodei and Chris Olah and Jacob Steinhardt and Paul Christiano and John Schulman and Dan Mané},
      year={2016},
      eprint={1606.06565},
      archivePrefix={arXiv},
      primaryClass={cs.AI}
}

@misc{hendrycks2018baseline,
      title={A Baseline for Detecting Misclassified and Out-of-Distribution Examples in Neural Networks}, 
      author={Dan Hendrycks and Kevin Gimpel},
      year={2018},
      eprint={1610.02136},
      archivePrefix={arXiv},
      primaryClass={cs.NE}
}

@INPROCEEDINGS{5206848,
  author={Deng, Jia and Dong, Wei and Socher, Richard and Li, Li-Jia and Kai Li and Li Fei-Fei},
  booktitle={2009 IEEE Conference on Computer Vision and Pattern Recognition}, 
  title={ImageNet: A large-scale hierarchical image database}, 
  year={2009},
  volume={},
  number={},
  pages={248-255},
  doi={10.1109/CVPR.2009.5206848}}

@article{Russakovsky2015ImageNetLS,
  title={ImageNet Large Scale Visual Recognition Challenge},
  author={Olga Russakovsky and Jia Deng and Hao Su and Jonathan Krause and Sanjeev Satheesh and Sean Ma and Zhiheng Huang and Andrej Karpathy and Aditya Khosla and Michael S. Bernstein and Alexander C. Berg and Li Fei-Fei},
  journal={International Journal of Computer Vision},
  year={2015},
  volume={115},
  pages={211-252}
}

@misc{hendrycks2017early,
      title={Early Methods for Detecting Adversarial Images}, 
      author={Dan Hendrycks and Kevin Gimpel},
      year={2017},
      eprint={1608.00530},
      archivePrefix={arXiv},
      primaryClass={cs.LG}
}

@misc{goodfellow2015explaining,
      title={Explaining and Harnessing Adversarial Examples}, 
      author={Ian J. Goodfellow and Jonathon Shlens and Christian Szegedy},
      year={2015},
      eprint={1412.6572},
      archivePrefix={arXiv},
      primaryClass={stat.ML}
}

@misc{kurakin2017adversarial,
      title={Adversarial Machine Learning at Scale}, 
      author={Alexey Kurakin and Ian Goodfellow and Samy Bengio},
      year={2017},
      eprint={1611.01236},
      archivePrefix={arXiv},
      primaryClass={cs.CV}
}

@misc{papernot2015limitations,
      title={The Limitations of Deep Learning in Adversarial Settings}, 
      author={Nicolas Papernot and Patrick McDaniel and Somesh Jha and Matt Fredrikson and Z. Berkay Celik and Ananthram Swami},
      year={2015},
      eprint={1511.07528},
      archivePrefix={arXiv},
      primaryClass={cs.CR}
}

@inproceedings{
guo2018countering,
title={Countering Adversarial Images using Input Transformations},
author={Chuan Guo and Mayank Rana and Moustapha Cisse and Laurens van der Maaten},
booktitle={International Conference on Learning Representations},
year={2018},
url={https://openreview.net/forum?id=SyJ7ClWCb},
}

@misc{gardner2016deep,
      title={Deep Manifold Traversal: Changing Labels with Convolutional Features}, 
      author={Jacob R. Gardner and Paul Upchurch and Matt J. Kusner and Yixuan Li and Kilian Q. Weinberger and Kavita Bala and John E. Hopcroft},
      year={2016},
      eprint={1511.06421},
      archivePrefix={arXiv},
      primaryClass={cs.LG}
}

@InProceedings{pmlr-v28-bengio13, title = {Better Mixing via Deep Representations}, author = {Yoshua Bengio and Gregoire Mesnil and Yann Dauphin and Salah Rifai}, booktitle = {Proceedings of the 30th International Conference on Machine Learning}, pages = {552--560}, year = {2013}, editor = {Sanjoy Dasgupta and David McAllester}, volume = {28}, number = {1}, series = {Proceedings of Machine Learning Research}, address = {Atlanta, Georgia, USA}, month = {6}, publisher = {PMLR}, url = {http://proceedings.mlr.press/v28/bengio13.html} }

@misc{carlini2017evaluating,
      title={Towards Evaluating the Robustness of Neural Networks}, 
      author={Nicholas Carlini and David Wagner},
      year={2017},
      eprint={1608.04644},
      archivePrefix={arXiv},
      primaryClass={cs.CR}
}

@misc{metzen2017detecting,
      title={On Detecting Adversarial Perturbations}, 
      author={Jan Hendrik Metzen and Tim Genewein and Volker Fischer and Bastian Bischoff},
      year={2017},
      eprint={1702.04267},
      archivePrefix={arXiv},
      primaryClass={stat.ML}
}

@article{DBLP:journals/corr/CarliniW17,
  author    = {Nicholas Carlini and
               David A. Wagner},
  title     = {Adversarial Examples Are Not Easily Detected: Bypassing Ten Detection
               Methods},
  journal   = {CoRR},
  volume    = {abs/1705.07263},
  year      = {2017},
  url       = {http://arxiv.org/abs/1705.07263},
  archivePrefix = {arXiv},
  eprint    = {1705.07263},
  bibsource = {dblp computer science bibliography, https://dblp.org}
}

@misc{hosseini2017blocking,
      title={Blocking Transferability of Adversarial Examples in Black-Box Learning Systems}, 
      author={Hossein Hosseini and Yize Chen and Sreeram Kannan and Baosen Zhang and Radha Poovendran},
      year={2017},
      eprint={1703.04318},
      archivePrefix={arXiv},
      primaryClass={cs.LG}
}

@article{Xu_2018,
   title={Feature Squeezing: Detecting Adversarial Examples in Deep Neural Networks},
   ISBN={1891562495},
   url={http://dx.doi.org/10.14722/ndss.2018.23198},
   DOI={10.14722/ndss.2018.23198},
   journal={Proceedings 2018 Network and Distributed System Security Symposium},
   publisher={Internet Society},
   author={Xu, Weilin and Evans, David and Qi, Yanjun},
   year={2018}
}

@article{JMLR:v9:vandermaaten08a,
  author  = {Laurens van der Maaten and Geoffrey Hinton},
  title   = {Visualizing Data using t-SNE},
  journal = {Journal of Machine Learning Research},
  year    = {2008},
  volume  = {9},
  number  = {86},
  pages   = {2579-2605},
  url     = {http://jmlr.org/papers/v9/vandermaaten08a.html}
}

@misc{papernot2017practical,
      title={Practical Black-Box Attacks against Machine Learning}, 
      author={Nicolas Papernot and Patrick McDaniel and Ian Goodfellow and Somesh Jha and Z. Berkay Celik and Ananthram Swami},
      year={2017},
      eprint={1602.02697},
      archivePrefix={arXiv},
      primaryClass={cs.CR}
}

@inproceedings{
buckman2018thermometer,
title={Thermometer Encoding: One Hot Way To Resist Adversarial Examples},
author={Jacob Buckman and Aurko Roy and Colin Raffel and Ian Goodfellow},
booktitle={International Conference on Learning Representations},
year={2018},
url={https://openreview.net/forum?id=S18Su--CW},
}

@misc{pang2018robust,
      title={Towards Robust Detection of Adversarial Examples}, 
      author={Tianyu Pang and Chao Du and Yinpeng Dong and Jun Zhu},
      year={2018},
      eprint={1706.00633},
      archivePrefix={arXiv},
      primaryClass={cs.LG}
}

@inbook{inbook,
author = {Carrara, Fabio and Becarelli, Rudy and Caldelli, Roberto and Falchi, Fabrizio and Amato, Giuseppe},
year = {2019},
month = {01},
pages = {313-327},
title = {Adversarial Examples Detection in Features Distance Spaces: Subvolume B},
isbn = {978-3-662-53906-4},
doi = {10.1007/978-3-030-11012-3_26}
}

@misc{samangouei2018defensegan,
      title={Defense-GAN: Protecting Classifiers Against Adversarial Attacks Using Generative Models}, 
      author={Pouya Samangouei and Maya Kabkab and Rama Chellappa},
      year={2018},
      eprint={1805.06605},
      archivePrefix={arXiv},
      primaryClass={cs.CV}
}

@article{goodfellow2014explaining,
  title={Explaining and harnessing adversarial examples},
  author={Goodfellow, Ian J and Shlens, Jonathon and Szegedy, Christian},
  journal={arXiv preprint arXiv:1412.6572},
  year={2014}
}

@misc{hwang2019puvae,
      title={PuVAE: A Variational Autoencoder to Purify Adversarial Examples}, 
      author={Uiwon Hwang and Jaewoo Park and Hyemi Jang and Sungroh Yoon and Nam Ik Cho},
      year={2019},
      eprint={1903.00585},
      archivePrefix={arXiv},
      primaryClass={cs.LG}
}

@misc{meng2017magnet,
      title={MagNet: a Two-Pronged Defense against Adversarial Examples}, 
      author={Dongyu Meng and Hao Chen},
      year={2017},
      eprint={1705.09064},
      archivePrefix={arXiv},
      primaryClass={cs.CR}
}

@misc{li2020deeprobust,
      title={DeepRobust: A PyTorch Library for Adversarial Attacks and Defenses}, 
      author={Yaxin Li and Wei Jin and Han Xu and Jiliang Tang},
      year={2020},
      eprint={2005.06149},
      archivePrefix={arXiv},
      primaryClass={cs.LG}
}

@inproceedings{carlini2017towards,
  title={Towards evaluating the robustness of neural networks},
  author={Carlini, Nicholas and Wagner, David},
  booktitle={IEEE symposium on security and privacy (S\&P)},
  pages={39--57},
  year={2017}
}

@article{akhtar2018threat,
  title={Threat of adversarial attacks on deep learning in computer vision: A survey},
  author={Akhtar, Naveed and Mian, Ajmal},
  journal={IEEE Access},
  volume={6},
  pages={14410--14430},
  year={2018}
}

@article{hendrycks2016early,
  title={Early methods for detecting adversarial images},
  author={Hendrycks, Dan and Gimpel, Kevin},
  journal={arXiv preprint arXiv:1608.00530},
  year={2016}
}

@inproceedings{li2017adversarial,
  title={Adversarial examples detection in deep networks with convolutional filter statistics},
  author={Li, Xin and Li, Fuxin},
  booktitle={Proceedings of the IEEE International Conference on Computer Vision},
  pages={5764--5772},
  year={2017}
}

@article{gong2017adversarial,
  title={Adversarial and clean data are not twins},
  author={Gong, Zhitao and Wang, Wenlu and Ku, Wei-Shinn},
  journal={arXiv preprint arXiv:1704.04960},
  year={2017}
}

@article{feinman2017detecting,
  title={Detecting adversarial samples from artifacts},
  author={Feinman, Reuben and Curtin, Ryan R and Shintre, Saurabh and Gardner, Andrew B},
  journal={arXiv preprint arXiv:1703.00410},
  year={2017}
}

@inproceedings{gao2021maximum,
  title={Maximum Mean Discrepancy Test is Aware of Adversarial Attacks},
  author={Gao, Ruize and Liu, Feng and Zhang, Jingfeng and Han, Bo and Liu, Tongliang and Niu, Gang and Sugiyama, Masashi},
  booktitle={International Conference on Machine Learning},
  pages={3564--3575},
  year={2021}
}

@inproceedings{Laine2019HighQualitySD,
  title={High-Quality Self-Supervised Deep Image Denoising},
  author={Samuli Laine and Tero Karras and Jaakko Lehtinen and Timo Aila},
  booktitle={NeurIPS},
  year={2019}
}

@INPROCEEDINGS{8954476,
  author={Raff, Edward and Sylvester, Jared and Forsyth, Steven and McLean, Mark},
  booktitle={2019 IEEE/CVF Conference on Computer Vision and Pattern Recognition (CVPR)}, 
  title={Barrage of Random Transforms for Adversarially Robust Defense}, 
  year={2019},
  volume={},
  number={},
  pages={6521-6530},
  doi={10.1109/CVPR.2019.00669}}

@article{song2017pixeldefend,
  title={Pixeldefend: Leveraging generative models to understand and defend against adversarial examples},
  author={Song, Yang and Kim, Taesup and Nowozin, Sebastian and Ermon, Stefano and Kushman, Nate},
  journal={arXiv preprint arXiv:1710.10766},
  year={2017}
}

@article{xu2017feature,
  title={Feature squeezing: Detecting adversarial examples in deep neural networks},
  author={Xu, Weilin and Evans, David and Qi, Yanjun},
  journal={arXiv preprint arXiv:1704.01155},
  year={2017}
}

@misc{moosavidezfooli2016deepfool,
      title={DeepFool: a simple and accurate method to fool deep neural networks}, 
      author={Seyed-Mohsen Moosavi-Dezfooli and Alhussein Fawzi and Pascal Frossard},
      year={2016},
      eprint={1511.04599},
      archivePrefix={arXiv},
      primaryClass={cs.LG}
}

@inproceedings{jha2018detecting,
  title={Detecting adversarial examples using data manifolds},
  author={Jha, Susmit and Jang, Uyeong and Jha, Somesh and Jalaian, Brian},
  booktitle={IEEE Military Communications Conference (MILCOM)},
  pages={547--552},
  year={2018}
}

@inproceedings{fidel2020explainability,
  title={When explainability meets adversarial learning: Detecting adversarial examples using shap signatures},
  author={Fidel, Gil and Bitton, Ron and Shabtai, Asaf},
  booktitle={International Joint Conference on Neural Networks (IJCNN)},
  pages={1--8},
  year={2020}
}

@article{lecun2010mnist,
  title={MNIST handwritten digit database},
  author={LeCun, Yann and Cortes, Corinna and Burges, CJ},
  journal={ATT Labs [Online]. Available: http://yann.lecun.com/exdb/mnist},
  volume={2},
  year={2010}
}

@TECHREPORT{Krizhevsky09learningmultiple,
    author = {Alex Krizhevsky},
    title = {Learning multiple layers of features from tiny images},
    institution = {University of Toronto},
    year = {2009}
}

@misc{Idelbayev18a,
  author       = "Yerlan Idelbayev",
  title        = "Proper {ResNet} Implementation for {CIFAR10/CIFAR100} in {PyTorch}"
}

@article{he2015deep,
      title={Deep Residual Learning for Image Recognition}, 
      author={Kaiming He and Xiangyu Zhang and Shaoqing Ren and Jian Sun},
      journal={arXiv preprint arXiv:1512.03385},
      year={2015}
}

@Book{EfroTibs93,
  Title                    = {An Introduction to the Bootstrap},
  Author                   = {Bradley Efron and Robert J. Tibshirani},
  Publisher                = {Chapman \& Hall/CRC},
  Year                     = {1993},

  Address                  = {Boca Raton, Florida, USA},
  Number                   = {57},
  Series                   = {Monographs on Statistics and Applied Probability}
}

@article{Kurakin2017AdversarialML,
  title={Adversarial Machine Learning at Scale},
  author={A. Kurakin and I. Goodfellow and Samy Bengio},
  journal={arXiv preprint arXiv:1611.01236},
  year={2017}
}
\addcontentsline{toc}{chapter}{References}
\end{document}